\documentclass{article}

\usepackage[main,final]{neurips_2026}

\usepackage{bm}

\usepackage[utf8]{inputenc} 
\usepackage[T1]{fontenc}    
\usepackage{hyperref}       
\usepackage{url}            
\usepackage{booktabs}       
\usepackage{amsfonts}       
\usepackage{nicefrac}       
\usepackage{microtype}      
\usepackage{xcolor}         
\usepackage{graphicx}
\usepackage{caption}
\usepackage{tcolorbox}
\usepackage{amsmath}
\usepackage{makecell}
\usepackage{diagbox}
\usepackage{wrapfig}
\usepackage{multirow}
\usepackage{algorithm}
\usepackage{algpseudocode}
\usepackage{xspace}
\usepackage{multicol}
\usepackage{bbm}
\newcommand{\ie}{\textit{i.e.}\xspace}

\newcommand{\ours}{GRGC}
\title{Revisiting On-policy Adversarial Black-Box Distillation: Calibrating Groupwise Reward Geometry for Effective Advantage Construction}

\author{Xiao Cui$^1$~~~ Mo Zhu$^{2*}$~~~Yulei Qin$^3$~~~ Wengang Zhou$^1$ ~~~ Yuze Wu$^{2}$~~~ Houqiang Li$^1$  \\
$^1$ University of Science and Technology of China 
\\ $^2$ Zhejiang University $^3$ Independent Researcher \\
{\tt\small cuixiao2001@mail.ustc.edu.cn, \{mozhu,wuyuze000\}@zju.edu.cn}
\\ 
{\tt\small qinyulei@sjtu.edu.cn, \{zhwg,lihq\}@ustc.edu.cn}}

\begin{document}

\maketitle
\makeatletter{\renewcommand*{\@makefnmark}{}
\footnotetext{*: Contributing equally with the first author.}
\footnotetext{Corresponding authors: Wengang Zhou and Houqiang Li.}\makeatother}

\begin{abstract}
Black-box distillation is a practical route for transferring capabilities from API-accessible large language models that expose only text outputs into smaller student models.
Recent on-policy adversarial methods such as 
GAD
improve over 
SeqKD by forming an adversarial loop between a critic and a student, 
where the critic provides rewards for GRPO-based student policy optimization over the student's sampled responses.
However, GRPO computes advantages from the within-group relative rewards of student samples for the same prompt, whereas the critic is trained primarily to distinguish teacher responses from student responses. This objective mismatch can produce reward groups with collapsed scale or fragile margins, leading to brittle grouped optimization signals.
We propose Groupwise Reward Geometry Conditioning (GRGC), a two-stage framework that improves advantage construction by shaping student-side reward groups during both critic training and policy optimization. To improve critic-side conditioning, Gaussian groupwise Optimal Transport calibration regularizes the critic during training to produce reward groups with non-collapsed spread and smooth rank-wise gaps by matching sorted prompt-wise rewards to group-centered Gaussian quantiles. 
Building on this conditioned reward geometry, policy-side group power modulation reshapes the prompt-wise reward groups before they are converted into advantages, preserving the critic-induced ordering while increasing optimization-relevant margin separability. 
Extensive experiments across diverse teachers, student model families and scales, and training datasets demonstrate the effectiveness of GRGC on both in-distribution and out-of-distribution evaluations, while introducing negligible overhead over GAD.
The code is available at \url{https://github.com/2018cx/GRGC}.
\end{abstract}

\section{Introduction}

Large language models (LLMs) have achieved strong performance across NLP tasks~\cite{zhang2026instruction,du2026survey,qin2026incentivizing}, but their scale leads to high computational and deployment costs~\cite{bedi2026holistic,niu2026tokenpowerbench,fang2026knowledge}. Knowledge distillation (KD) mitigates this issue by transferring capabilities from large teacher models to smaller students~\cite{zhang2026askd,wang2026bridging,vuong2026mcw}. Most existing KD methods assume a \emph{white-box} setting, where internal teacher signals such as logits and hidden states are accessible, enabling objectives based on distribution and representation alignment~\cite{sanh2019distilbert,gu2024minillm,agarwal2024onpolicy}. However, modern frontier LLMs are typically accessible only via text-generation APIs~\cite{gpt5,comanici2025gemini,cui2024exploring,sun2024exploiting}, which restricts supervision to sampled outputs and defines the \emph{black-box distillation} setting. In this regime, standard white-box objectives become inapplicable, making distillation fundamentally more challenging and necessitating new learning paradigms~\cite{jiang2023lion,gad}.

Early black-box distillation methods~\cite{taori2023alpaca,vicuna,deepseekr1} typically adopt sequence-level knowledge distillation (SeqKD)~\cite{seqkd}, which treats teacher-generated responses as supervised targets for direct student training. SeqKD is simple and scalable, but it is fundamentally off-policy: the student is trained on teacher trajectories and then tested on its own rollouts, leading to exposure bias and a persistent train-test mismatch. This limitation has motivated recent moves toward on-policy black-box distillation~\cite{xiong2026ovd,gad,jiang2023lion}. Among these directions, on-policy adversarial  distillation~\cite{gad,wang2026prism} is particularly attractive because it replaces continual online teacher scoring with a learned critic, making training more cost-effective than approaches that repeatedly query the teacher during policy optimization. This practical advantage, however, exposes a sharper critic-to-advantage interface problem. The critic is trained to assign higher scores to teacher responses than to student responses, but GRPO updates the student from prompt-wise advantages computed only over the student's own sampled response group. Because the critic objective does not explicitly constrain the gaps, ordering robustness, or scale of this student-only reward group, these geometric properties can deteriorate even when teacher--student discrimination improves. Our analysis and propositions show that this deterioration directly affects advantage construction: low within-group dispersion makes the student-side ordering fragile under critic noise, while weak group scale amplifies noise sensitivity in the downstream advantage signal.


We call this 
groupwise reward geometry mismatch. To address it, we propose Groupwise Reward Geometry Conditioning (GRGC), 
which intervenes at two coupled stages: critic-side formation of student reward groups and their policy-side use in GRPO.
Gaussian groupwise Optimal Transport (OT) calibration acts during critic training, at the source where critic scores form the raw student-side reward group.
It uses an efficient OT-based objective to align the sorted rewards in each prompt group with group-centered Gaussian quantiles, chosen for their symmetric, moderate rank-gap profile, thereby inducing a non-degenerate spread, a stable scale, and a smooth ordered gap profile.
This geometry is optimization-relevant: it makes within-group comparisons less prone to ranking flips and less sensitive to perturbations.
Policy-side group modulation then acts on this conditioned reward geometry.
After critic-side geometry calibration reduces collapse and scale drift, centering and scaling each reward group yields standardized scores with more reliable relative positions.
In this regime, our analysis shows that 
the signed power transform preserves the critic-induced ordering and enlarges informative within-group margins under bounded perturbations, yielding more separable reward groups for GRPO-based student policy training.
Our propositions further make this optimization link explicit: in a grouped softmax surrogate, advantage gaps exactly determine one-step pairwise log-odds shifts, while advantage errors induce bounded distortion in the clipped GRPO surrogate.

\begin{figure}[t]
\centering
\includegraphics[width=0.98\linewidth]{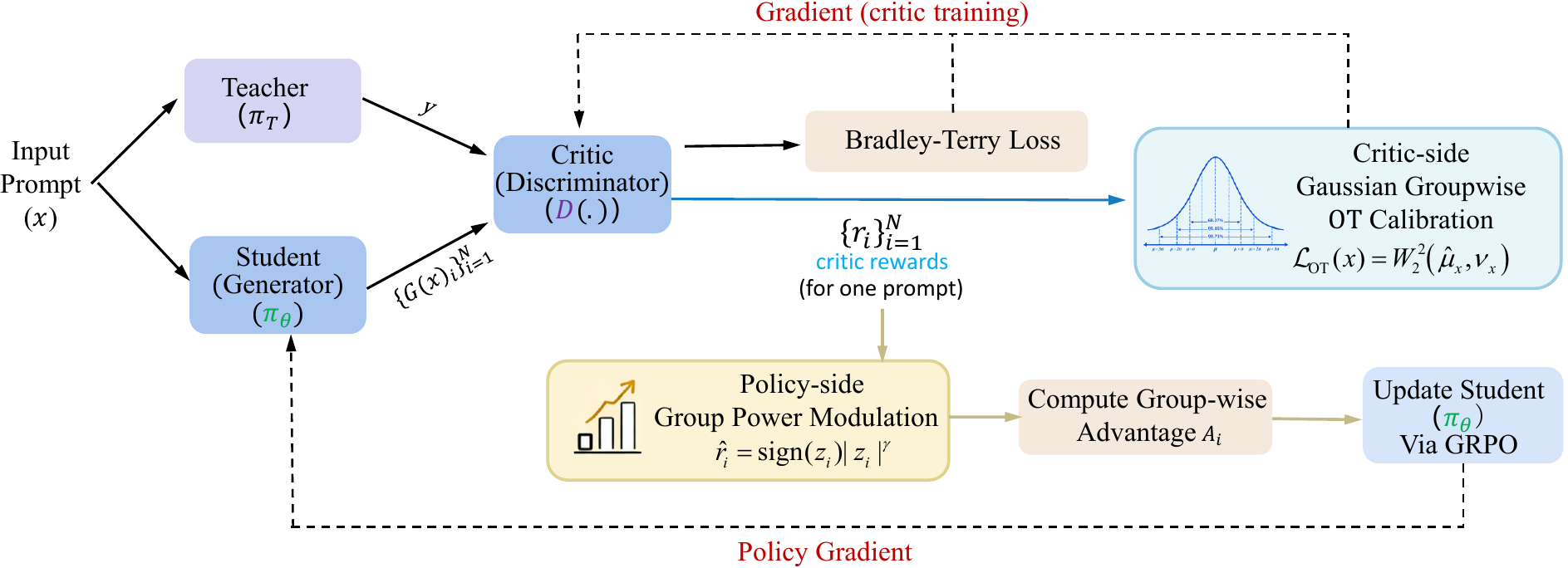}
\vspace{-0.6em}
\caption{\textbf{Overview of GRGC.} For each prompt $x$, the student samples a response group $\{G(x)_i\}_{i=1}^N$, and the critic assigns raw sequence-level rewards $\{r_i\}_{i=1}^N$. CGC regularizes prompt-group reward geometry during critic training, while PGM transforms the conditioned rewards into $\{\hat r_i\}_{i=1}^N$ before the final prompt-wise advantages $\{A_i\}_{i=1}^N$ are computed for policy optimization.}
\vspace{-1.4em}
\label{fig:frameworkmain}
\end{figure}

Our overall pipeline is illustrated in Figure~\ref{fig:frameworkmain}, and our contributions are summarized as follows:
\begin{itemize}
\item We identify the groupwise reward-geometry mismatch in on-policy adversarial black-box distillation, where the critic can learn effective teacher--student discrimination while 
producing student-side reward groups that are poorly conditioned for GRPO advantage estimation.
\item We propose GRGC, a lightweight framework for effective advantage construction: critic-side geometry calibration first stabilizes prompt-wise reward groups, and policy-side group modulation then amplifies informative margins on the conditioned signal. 
\item Experiments across diverse teachers, student model families and scales, and training datasets show consistent gains on in-distribution and out-of-distribution benchmarks, supported by different auto-judging models, evaluation metrics, and human preference evaluation.
\end{itemize}

\section{Related Works}


Teacher-based LLM distillation methods are broadly divided into white-box and black-box distillation.

\subsection{White-box Knowledge Distillation of LLMs}
White-box knowledge distillation leverages full access to the teacher's internal states, such as token-level probabilities and hidden representations, to guide student learning \cite{hinton2015distilling, gou2021knowledge}. 
Traditional methods mainly align output distributions with forward Kullback--Leibler (KL) divergence~\cite{sanh2019distilbert,gu2024minillm}, but some studies show that forward KL encourages mode-covering behavior, which can overburden limited-capacity students~\cite{gu2024minillm,wu2024rethinking}. 
To address this, 
MiniLLM \cite{gu2024minillm} and GKD \cite{agarwal2024onpolicy} use reverse KL divergence to promote mode-seeking behavior, ensuring the student focuses on the teacher's high-confidence reasoning paths. Furthermore, structural distillation through intermediate features, such as hidden states and attention maps in TinyBERT \cite{jiao2020tinybert}, has proven essential for deep architectural imitation.
Beyond direct distribution matching, MMKD~\cite{jia2024adversarial} formulates distillation as adversarial action-value moment matching.
Recent studies have explored cross-tokenizer knowledge distillation to overcome conventional KD's reliance on identical teacher--student tokenizers and enable cross-family model transfer.
ULD \cite{boizard2024towards} addresses vocabulary misalignment by using the Wasserstein distance to align disparate token spaces. DSKD \cite{zhang2024dualspace} unifies prediction spaces through bidirectional linear projections and an Exact Token Alignment algorithm. 
Complementary approaches like ALM \cite{minixhofer2025universal} and DistillMoE \cite{minixhofer2025distillmoe} explore approximate likelihood matching and mixture-of-experts strategies to further stabilize cross-tokenizer knowledge transfer. 
However, white-box distillation is impractical for state-of-the-art models such as GPT-5~\cite{gpt5} and Doubao-Seed~\cite{Seed2}, whose internal states are typically inaccessible.

\subsection{Black-box Knowledge Distillation of LLMs}
Black-box distillation uses only observable teacher outputs rather than internal teacher signals.
The dominant paradigm is 
SeqKD~\cite{seqkd}, which treats teacher responses as supervised labels and has supported many open-source instruction-tuned models~\cite{taori2023alpaca,vicuna,peng2023instruction,zhou2023lima}. However, SeqKD is inherently off-policy and therefore vulnerable to exposure bias and distribution shift~\cite{gudibande2024false,setiawan2024accurate}. To enrich supervision under this paradigm, later works have explored richer reasoning traces~\cite{mukherjee2023orca,hsieh2023distilling,openthoughts,limo,deepseekr1,s1}. 
Motivated by the success of white-box on-policy distillation~\cite{gu2024minillm,agarwal2024onpolicy,ko2024distillm}, recent works have extended this idea to black-box distillation~\cite{xiong2026ovd,gad}.
OVD~\cite{xiong2026ovd} obtains on-policy supervision by repeatedly querying the black-box teacher for verbal scores on student-generated trajectories.
While effective, this teacher-in-the-loop design incurs substantial inference cost.
GAD~\cite{gad} instantiates adversarial 
distillation by introducing a critic model that is trained alternately with the student under a GAN-style minimax objective.
The critic provides online rewards for student-generated responses, which are converted into prompt-wise advantages for GRPO-based policy optimization, while offline-generated teacher responses can be reused throughout training to avoid continual teacher querying.
However, replacing online teacher supervision with an online critic shifts the key challenge to the critic-to-advantage interface. 
We address this bottleneck by calibrating groupwise reward geometry. 

\section{Methods}

\subsection{Problem Setup}
We study black-box distillation from an API-accessible teacher policy $\pi_T$ to a student policy $\pi_\theta$. Given a prompt $x$, the teacher produces a response $y \sim \pi_T(\cdot \mid x)$. The objective is to optimize $\pi_\theta$ to approximate $\pi_T$ using only teacher responses, without access to the teacher's internal signals.

\subsection{Revisiting On-policy Adversarial Black-Box Distillation}
To reduce the risk of reward hacking against a fixed reward model, on-policy adversarial black-box distillation~\cite{gad,wang2026prism} instantiates black-box distillation through an alternating student--critic training loop within a grouped policy optimization framework. For each prompt $x$, the student policy $\pi_\theta$ samples a response group $\{G(x)_i\}_{i=1}^N$, where $N$ is the group size. 
The critic, serving as an online reward model, assigns sequence-level scores to both 
student responses and the teacher response $y$; these scores are used 
to form the critic's adversarial training objective. 
The student-side 
scores are then used as rewards to construct 
within-group 
advantages for GRPO-based optimization of $\pi_\theta$.

GAD~\cite{gad} trains the critic model with a Bradley--Terry-style objective:
\begin{equation}
\mathcal{L}_{\mathrm{BT}}(x)
=
\frac{1}{N}\sum_{i=1}^N - \log \sigma\!\left(D(y) - D(G(x)_i)\right),
\end{equation}
where $D(\cdot)$ is the critic score and $\sigma(\cdot)$ is the logistic sigmoid. The core mismatch is especially relevant in adversarial black-box distillation, where the critic is updated online rather than calibrated as a general preference evaluator. Its BT objective enforces teacher--student discrimination, which can be achieved without within-prompt separability among student responses. GRPO, however, builds the student update from prompt-wise advantages derived from student-side rewards, making the update sensitive to their within-group reward geometry.
\par\noindent\textbf{Proposition 1.} Fix the teacher score $D(y)$ and let $\mu_x=\frac{1}{N}\sum_{i=1}^N D(G(x)_i)$ be the mean student score. Then strict convexity and Jensen's inequality~\cite{mcshane1937jensen} give
\begin{equation}
\mathcal{L}_{\mathrm{BT}}(x)
\ge
\log\!\left(1+\exp(\mu_x-D(y))\right),
\end{equation}
with equality if and only if $D(G(x)_1)=\cdots=D(G(x)_N)=\mu_x$. Thus, on every fixed-mean slice, the BT objective uniquely favors zero within-group dispersion. The proof is given as Proposition~1 in Appendix~\ref{a2}.

\begin{figure}[t]
\centering
\includegraphics[width=0.98\linewidth]{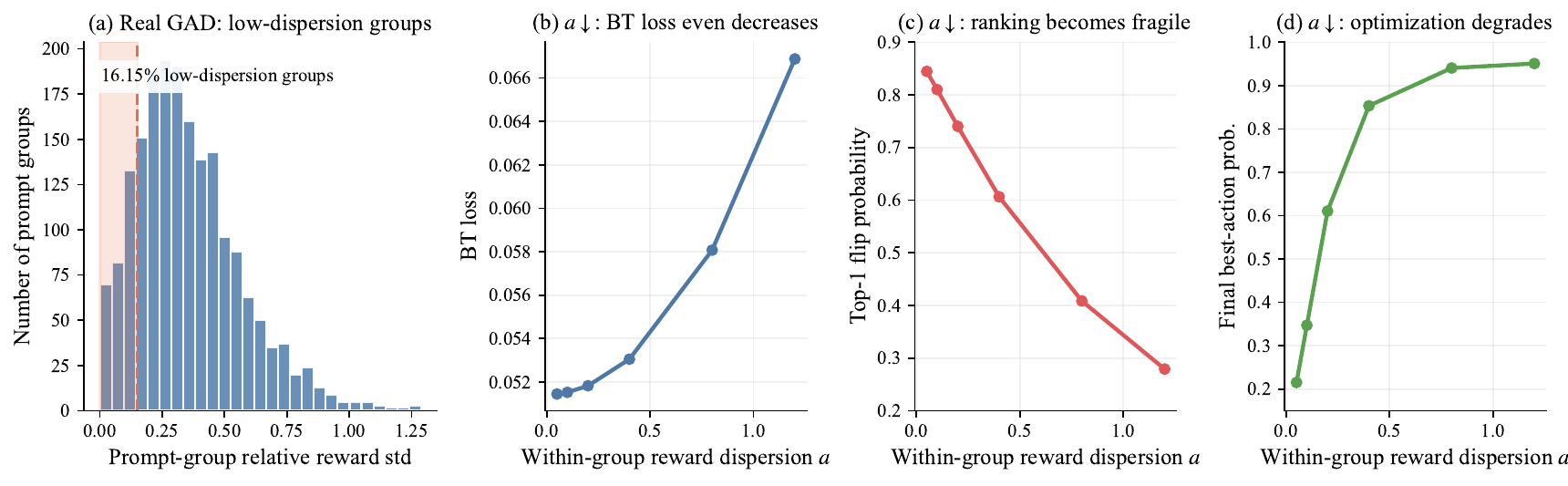}
\vspace{-0.2em}
\caption{\textbf{Why BT-style adversarial critic training alone is insufficient for advantage construction.} (a) Real GAD training often yields low-dispersion prompt groups. The x-axis denotes the group reward standard deviation normalized by the global reward standard deviation at the same step. (b)--(d) Toy grouped-bandit analysis. We fix latent utility scores $u_i$ to define the within-group preference order and set reward $r_i^{(a)}=a\,u_i$, so that varying $a$ changes only the within-group reward dispersion. Smaller $a$ means a more collapsed reward group. As $a$ decreases, BT loss does not increase and may even decrease, while ranking becomes more noise-sensitive and grouped optimization deteriorates.
}
\vspace{-0.5em}
\label{fig:gad_pathology}
\end{figure}

\begin{wraptable}{r}{0.5\textwidth}
\centering
\vspace{-1.1em}
\caption{Low-dispersion incidence over the complete 1500-step Qwen2.5-3B/GPT-5/LMSYS trajectories. Each stage contains 64000 groups per method.}
\label{tab:full_trajectory_dispersion}
\scriptsize
\setlength{\tabcolsep}{2.3pt}
\resizebox{\linewidth}{!}{%
\begin{tabular}{lcc|cc}
\toprule
& \multicolumn{2}{c|}{Rel. std $<0.15$} & \multicolumn{2}{c}{Rel. std $<0.25$} \\
Stage & GAD & GRGC & GAD & GRGC \\
\midrule
Early (500 steps)  & 8.57\% & \textbf{1.28\%} & 15.73\% & \textbf{1.28\%} \\
Middle (500 steps) & 8.69\% & \textbf{1.90\%} & 16.04\% & \textbf{1.90\%} \\
Late (500 steps)   & 8.48\% & \textbf{1.86\%} & 14.73\% & \textbf{1.86\%} \\
\bottomrule
\end{tabular}}
\vspace{-1.5em}
\end{wraptable}

Figure~\ref{fig:gad_pathology} provides visual evidence for this mismatch. Figure~\ref{fig:gad_pathology}(a) shows that real GAD training produces a non-negligible fraction of low-dispersion prompt groups, indicating that the emitted reward geometry can already be weak in practice. 
In a toy grouped-bandit construction, we adjust the within-group reward dispersion while leaving the teacher--student preference direction essentially unchanged. 
Figure~\ref{fig:gad_pathology}(b) shows that this deterioration in geometry does not worsen the BT loss and can even reduce it, consistent with Proposition~1: the critic objective does not protect student-side dispersion and, at fixed mean, strictly favors collapse. However, once the reward group becomes too flat, the within-group ordering becomes much more fragile under noise, as shown in Figure~\ref{fig:gad_pathology}(c). This increased fragility then directly degrades grouped optimization, as shown in Figure~\ref{fig:gad_pathology}(d).

Table~\ref{tab:full_trajectory_dispersion} extends Figure~\ref{fig:gad_pathology}(a) to the complete run. GAD remains near $8.5\%$/$15\%$ at the $0.15$/$0.25$ thresholds throughout training, whereas GRGC never exceeds $1.90\%$. The equal GRGC columns are exact: every flagged group contains eight identical responses, while its minimum nonzero relative std is $0.8105$. Thus, GAD's pathology persists beyond initialization or a selected interval.





\noindent\textbf{Proposition 2.} Let $\Delta_{\min}(r)=\min_{i\ne j,r_i\ne r_j}|r_i-r_j|$ denote the smallest nonzero group margin. Under bounded perturbations $|\eta_i|\le \eta$, the ordering is guaranteed to be preserved if and only if
$\Delta_{\min}(r)>2\eta$.
Thus low dispersion directly makes the grouped comparison structure fragile.

\noindent\textbf{Proposition 3.} Let $z=(z_i)_{i=1}^N$ be the prompt-wise normalized reward vector. For a perturbed group
$\tilde r=r+\zeta$, let $\tilde z$ be its normalized vector. Then
$
\|\tilde z-z\|_2
=
\mathcal{O}\!\left(\frac{\|\zeta\|_2}{\sigma_x+\varepsilon}\right).
$
Weak group scale amplifies critic noise after grouped normalization and makes the downstream advantage signal more sensitive.

They correspond to Propositions~2 and 3 in Appendix~\ref{a2}, where the full statements and proofs are given. 
Together with Figure~\ref{fig:gad_pathology}(b), these facts explain why teacher--student discrimination alone is insufficient: the BT objective can remain favorable while the student-side reward geometry becomes too flat, noise-sensitive, and poorly conditioned for advantage construction.

\begin{wrapfigure}{r}{0.7\textwidth}
    \centering
    \vspace{-1em}
    \includegraphics[width=\linewidth]{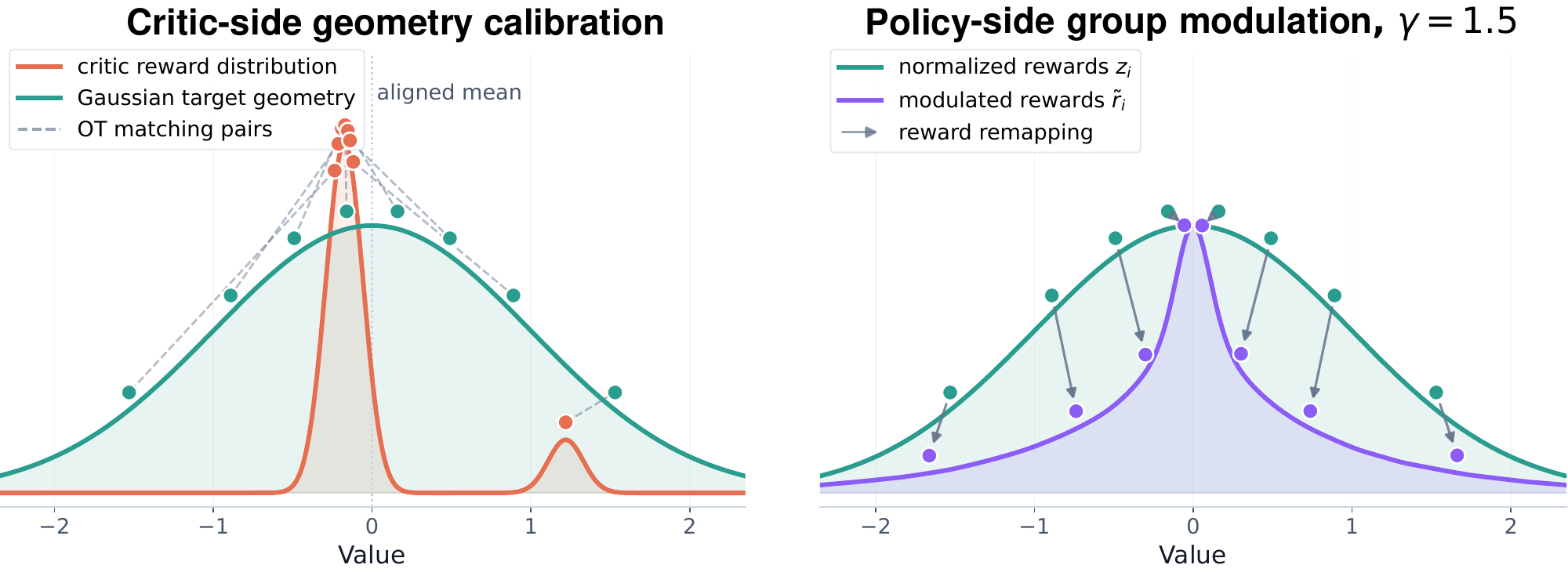}
    \caption{
    \textbf{Mechanism view of GRGC.} Critic-side Gaussian OT calibration aligns 
prompt-group rewards with mean-centered Gaussian quantiles, promoting non-collapsed scale and
smooth rank-wise gaps. Policy-side group power modulation remaps group
rewards to enlarge informative margins while preserving the critic-induced ordering.
    }
    \label{fig:framework}
    \vspace{-0.8em}
\end{wrapfigure}


\subsection{Calibrating Groupwise Reward Geometry for Effective Advantage Construction}


The geometry mismatch above motivates an intervention at the critic-to-advantage interface. To address this, we explicitly condition prompt-wise reward group geometry for effective advantage construction. 
The preceding analysis shows that student-side critic rewards must avoid collapse, maintain stable within-group scale, and preserve sufficient separation for reliable advantage estimation. Additionally, prior studies~\cite{wang2026kalmanfilterenhancedgrpo} emphasize the importance of stable within-group variance for this estimation, both of which motivate the need for critic-side geometry calibration.
While geometry calibration improves conditioning, it does not actively sharpen the informative margins needed for early grouped policy separation. To address this, we introduce policy-side group modulation, which amplifies useful grouped margins before deriving the final advantages.
GRGC, therefore, consists of two complementary components:
\textbf{(1) Critic-side geometry calibration (CGC):} During critic training, an OT objective regularizes reward groups toward well-separated, non-collapsed, and variance-stable geometry.
\textbf{(2) Policy-side group modulation (PGM):} During policy optimization, a group power transform reshapes critic rewards before the final prompt-wise advantages are computed for GRPO.
\paragraph{Gaussian Groupwise OT Calibration for Critic Training}
During critic training, we regularize each prompt-wise student reward group toward a reference geometry. The goal is to impose an optimization-friendly within-group structure: the group should avoid collapse, maintain a controlled spread, and provide smoothly varying rank-wise gaps. 
We use a Gaussian quantile template to provide balanced geometry within a symmetric location-scale family, preventing rigid spacing and excessive tail emphasis, and producing reward groups that are suitable for subsequent advantage construction.


For each prompt $x$, 
let $\hat{\mu}_x = \frac{1}{N} \sum_{i=1}^N \delta_{D(G(x)_i)}$ be the empirical student reward distribution. 
We introduce a Gaussian reference distribution $\nu_x = \mathcal{N}(\mu_x, 1)$, where $\mu_x = \frac{1}{N} \sum_{j=1}^N D(G(x)_j)$. By centering the reference at the current group mean, the calibration constrains only the centered shape and scale of the group. 
The critic-side calibration objective is defined as the squared 2-Wasserstein distance between the empirical distribution and the reference:
\begin{equation}
\mathcal{L}_{\mathrm{OT}}(x) = W_2^2(\hat{\mu}_x, \nu_x).
\end{equation}
In the one-dimensional case, the exact $W_2^2$ distance between a discrete $\hat{\mu}_x$ and a continuous $\nu_x$ theoretically involves an integral over the Gaussian quantile function. To ensure computational efficiency, we employ a discrete approximation of $\nu_x$ by sampling $N$ equally weighted quantiles. Let $r_{(1)} \le r_{(2)} \le \cdots \le r_{(N)}$ be the sorted rewards, and define the target Gaussian quantiles as:
\begin{equation}
q_j=\Phi^{-1}\!\left(\frac{j-0.5}{N}\right),
\qquad
t_j=\mu_x+q_j,
\quad j=1,\dots,N.
\end{equation}
In practice, we compute the Optimal Transport loss between  $\hat{\mu}_x$ and 
the equally weighted discrete reference measure
$\nu_x^{(N)} = \frac{1}{N} \sum_{j=1}^N \delta_{t_j}$, which admits the following closed-form expression~\cite{peyre2019computational}:
\begin{equation}
\mathcal{L}_{\mathrm{OT}}(x) =  W_2^2(\hat{\mu}_x, \nu_x^{(N)})=\frac{1}{N} \sum_{j=1}^N \bigl( r_{(j)} - t_j \bigr)^2.
\end{equation}
The group mean is the unique OT-optimal translation of this centered template. Specifically, Proposition~4.3 in Appendix~\ref{a2} proves that the cost for a translated template $a+q_j$ decomposes as
\begin{equation}
\mathcal{C}_x(a)=\mathcal{C}_x(\mu_x)+(a-\mu_x)^2,
\end{equation}
and, writing the group loss as a function of its reward vector, that $\mathcal{L}_{\mathrm{OT}}(r+c\mathbf{1})=\mathcal{L}_{\mathrm{OT}}(r)$. CGC therefore constrains within-group geometry while preserving common-score translation symmetry.
Accordingly, the critic training objective becomes:
\begin{equation}
\mathcal{L}_{\mathrm{critic}}(x)=\mathcal{L}_{\mathrm{BT}}(x)+\lambda_{\mathrm{OT}}\mathcal{L}_{\mathrm{OT}}(x).
\end{equation}

\noindent\textbf{Proposition 4.} Let $\sigma_x$ denote the empirical standard deviation of the centered reward group and $\sigma_q$ the standard deviation of the centered Gaussian target quantiles. Then
\begin{equation}
\mathcal{L}_{\mathrm{OT}}(x)\ge (\sigma_x-\sigma_q)^2.
\end{equation}
Hence any collapsed or severely under-dispersed prompt group necessarily incurs a large OT penalty.

Proposition~4 corresponds to Proposition~4.1 in Appendix~\ref{a2} and shows that CGC penalizes collapse and scale mismatch. Proposition~4.2 shows that OT penalizes ordered shape mismatch even when variance is matched, Proposition~4.3 establishes optimal mean anchoring and translation invariance, and Proposition~10 shows that Gaussian quantiles induce a more suitable interior-to-tail gap profile than linear or heavier-tailed alternatives. These analyses show that CGC is not only a variance regularizer, but a group-centered conditioner of the full ordered reward geometry.

\paragraph{Group Power Transform for Policy Optimization}
Critic-side calibration encourages the critic to emit stable, non-collapsed, and structurally regular reward groups, providing a well-conditioned geometry for grouped optimization. Building on this calibrated geometry, we apply a group power transform before final prompt-wise advantage computation to convert useful within-group differences into sufficiently decisive advantage gaps while preserving the critic-induced ordering.
By retaining relative magnitude information and enlarging informative margins with a controlled polynomial gain, this transform makes the resulting reward groups more responsive to grouped optimization.

For each prompt $x$, let $\{r_i\}_{i=1}^N$ denote the corresponding critic reward group. The signed power transform is symmetric around zero; therefore, centering the reward group provides a natural reference point from which positive and negative deviations can be amplified in a balanced manner:
\begin{equation}
z_i = \frac{r_i - \mu_x}{\sigma_x + \varepsilon},
\end{equation}
where $\mu_x$ and $\sigma_x$ are the mean and standard deviation of $\{r_i\}_{i=1}^N$. We then apply the group transform:
\begin{equation}
\hat{r}_i = \mathrm{sign}(z_i)\,|z_i|^\gamma, \quad \gamma > 1.
\label{grouppowerequation}
\end{equation}
GRPO subsequently derives prompt-wise advantages from the transformed rewards $\{\hat r_i\}_{i=1}^N$:
\begin{equation}
A_i = \frac{\hat{r}_i - \mathrm{mean}\!\left(\{\hat{r}_j\}_{j=1}^N\right)}
{\mathrm{std}\!\left(\{\hat{r}_j\}_{j=1}^N\right)+\varepsilon}.
\end{equation}

\noindent\textbf{Proposition 5.} 
Let $\hat r_i=f_\gamma(z_i)$ as defined in Eq.~\eqref{grouppowerequation}.
The transform $f_\gamma$ is strictly monotone and therefore preserves the critic-implied ordering.
For any same-sign pair with $\min(|z_i|,|z_j|)\ge \rho>0$,
\begin{equation}
|\hat r_i-\hat r_j| \ge \gamma \rho^{\gamma-1}|z_i-z_j|.
\end{equation}
Hence, already-informative within-group gaps are enlarged before GRPO processing.

Proposition~5 shows that PGM preserves ordering while enlarging already informative gaps. The proof is given in Appendix~\ref{a2}. Proposition~11 in the appendix formalizes the structural requirements for a pre-advantage transformation: it should preserve order, treat both sides symmetrically, apply a scale-consistent gain rule, and introduce no extra shape parameters. Under these requirements, the signed power form is uniquely determined among common transforms. Proposition~6 in the appendix further shows that the enlarged margins remain valid under bounded noise in the informative regime.

\paragraph{From advantage gaps to policy separation.}
The two components above define the mechanism chain of GRGC: CGC improves the geometry of rewards emitted by the critic, and PGM turns that geometry into more separable transformed rewards. In the grouped softmax surrogate analyzed in Appendix~\ref{a2}, let $\pi_i(\theta)$ and $\pi_j(\theta)$ denote the policy probabilities assigned to candidates $i$ and $j$ within the same prompt group. Proposition~7 shows that one update changes their 
pairwise log-odds by
\begin{equation}
\log\frac{\pi_i(\theta^+)}{\pi_j(\theta^+)}
-
\log\frac{\pi_i(\theta)}{\pi_j(\theta)}
=
\eta_{\mathrm{pg}}(A_i-A_j).
\end{equation}
Thus, reward geometry determines transformed reward separability, which determines the advantage gaps driving one-step policy separation in the grouped softmax surrogate. Proposition~8 shows that misranked constructed advantages can make this movement fail to improve the desired pairwise preference, or even move in the wrong direction. Proposition~9 connects this mechanism to clipped GRPO: errors in the constructed advantages induce bounded distortion of the clipped surrogate, so improving advantage conditioning and separability matter for the actual policy objective.

\begin{table*}[t]
\caption{
Automatic evaluation results for models distilled from GPT-5-Chat and trained on the LMSYS-Chat training set. We report the averaged Qwen2.5-72B evaluation score and win rate.
}
\vspace{-0.7em}
\centering
\resizebox{\textwidth}{!}{
\begin{tabular}{cc|cc|cc|cc|cc}
\toprule
\multirow{2}{*}{Model} & \multirow{2}{*}{Method}
& \multicolumn{2}{c|}{LMSYS}
& \multicolumn{2}{c|}{Dolly}
& \multicolumn{2}{c|}{SelfInst}
& \multicolumn{2}{c}{Vicuna} \\
&
& Score & Win
& Score & Win
& Score & Win
& Score & Win \\
\midrule
GPT-5-Chat & Teacher
& 51.21 & 53.2\%
& 49.57 & 49.4\%
& 49.71 & 49.2\%
& 50.27 & 55.0\% \\
\midrule

\multirow{4}{*}{Qwen2.5-3B-Instruct}
& Before Distill.
& 45.79 & 11.9\%
& 44.93 & 4.0\%
& 46.56 & 12.8\%
& 47.85 & 3.8\% \\
& SeqKD
& 47.06 & 18.4\%
& 45.62 & 7.2\%
& 46.83 & 16.9\%
& 48.25 & 20.0\% \\
& GAD
& 48.44 & 25.1\%
& 46.19 & 8.2\%
& 47.33 & 16.9\%
& 48.76 & 28.8\% \\
& \textbf{\ours}
& \textbf{50.19} & \textbf{45.3\%}
& \textbf{47.40} & \textbf{26.6\%}
& \textbf{48.99} & \textbf{30.6\%}
& \textbf{50.48} & \textbf{38.8\%} \\
\midrule

\multirow{4}{*}{Qwen2.5-1.5B-Instruct}
& Before Distill.
& 41.93 & 4.8\%
& 39.78 & 0.6\%
& 41.06 & 5.0\%
& 43.24 & 0.0\% \\
& SeqKD
& 45.43 & 18.8\%
& 42.91 & 4.6\%
& 44.34 & 11.2\%
& 47.41 & 11.3\% \\
& GAD
& 46.05 & 15.7\%
& 42.97 & 7.6\%
& 45.27 & 9.9\%
& 47.78 & 12.5\% \\
& \textbf{\ours}
& \textbf{47.36} & \textbf{26.1\%}
& \textbf{45.16} & \textbf{18.6\%}
& \textbf{47.42} & \textbf{20.2\%}
& \textbf{48.81} & \textbf{27.5\%} \\
\midrule

\multirow{4}{*}{Llama-3.2-3B-Instruct}
& Before Distill.
& 44.46 & 13.8\%
& 45.72 & 8.0\%
& 47.25 & 16.5\%
& 47.98 & 5.0\% \\
& SeqKD
& 45.99 & 17.3\%
& 47.19 & 11.2\%
& 47.38 & 16.5\%
& 48.40 & 22.5\% \\
& GAD
& 46.51 & 15.9\%
& 47.67 & 11.6\%
& 48.34 & 21.5\%
& 48.52 & 21.3\% \\
& \textbf{\ours}
& \textbf{49.07} & \textbf{40.1\%}
& \textbf{48.91} & \textbf{33.4\%}
& \textbf{49.28} & \textbf{30.2\%}
& \textbf{49.57} & \textbf{26.3\%} \\
\midrule

\multirow{4}{*}{Llama-3.2-1B-Instruct}
& Before Distill.
& 39.50 & 4.4\%
& 42.04 & 2.8\%
& 41.85 & 7.0\%
& 47.00 & 3.8\% \\
& SeqKD
& 43.04 & 11.9\%
& 42.30 & 5.2\%
& 43.26 & 10.3\%
& 46.89 & 7.5\% \\
& GAD
& 42.57 & 8.8\%
& 42.48 & 5.0\%
& 42.93 & 9.9\%
& 47.08 & 11.3\% \\
& \textbf{\ours}
& \textbf{44.68} & \textbf{18.2\%}
& \textbf{43.84} & \textbf{14.8\%}
& \textbf{44.82} & \textbf{12.8\%}
& \textbf{47.54} & \textbf{20.0\%} \\
\bottomrule
\end{tabular}
}
\label{tab:auto1}
\vspace{-0.3em}
\end{table*}

\begin{table}[t]
\caption{Automatic evaluation results for models distilled from Doubao-Seed-2.0 and trained on the LMSYS-Chat training set. We report the averaged Qwen2.5-72B evaluation score and win rate.}
\centering
\resizebox{\linewidth}{!}{
\begin{tabular}{cc|cc|cc|cc|cc}
\toprule
\multirow{2}{*}{Model} & \multirow{2}{*}{Method}
& \multicolumn{2}{c|}{LMSYS}
& \multicolumn{2}{c|}{Dolly}
& \multicolumn{2}{c|}{SelfInst}
& \multicolumn{2}{c}{Vicuna} \\
& 
& Score & Win
& Score & Win
& Score & Win
& Score & Win \\
\midrule
Doubao-Seed-2.0 & Teacher
& 53.23 & 82.5\%
& 53.14 & 75.8\%
& 52.70 & 74.0\%
& 53.20 & 93.8\% \\
\midrule

\multirow{4}{*}{Qwen2.5-3B-Instruct}
& Before Distill.
& 45.79 & 11.9\%
& 44.93 & 4.0\%
& 46.56 & 12.8\%
& 47.85 & 3.8\% \\
& SeqKD
& 46.95 & 25.3\%
& 45.96 & 9.6\%
& 46.87 & 16.9\%
& 47.81 & 15.0\% \\
& GAD
& 48.16 & 42.2\%
& 46.86 & 37.0\%
& 47.78 & 38.0\%
& 49.84 & 60.0\% \\
& \textbf{\ours}
& \bf 49.49 & \bf 49.9\%
& \bf 47.92 & \bf 48.0\%
& \bf 48.75 & \bf 43.0\%
& \bf 51.23 & \bf 68.8\% \\
\midrule

\multirow{4}{*}{Qwen2.5-1.5B-Instruct}
& Before Distill.
& 41.93 & 4.8\%
& 39.78 & 0.6\%
& 41.06 & 5.0\%
& 43.24 & 0.0\% \\
& SeqKD
& 44.65 & 17.7\% 
& 43.53 & 6.0\%
& 43.70 & 12.8\%
& 47.48 & 15.0\% \\
& GAD
& 45.13 & 27.1\%
& 43.23 & 24.8 \%
& 46.15 & 27.3\%
& 47.98 & 37.5\% \\
& \textbf{\ours}
& \bf 47.28 & \bf 30.9\%
&  \bf 46.40 & \bf29.2\%
&  \bf 46.94  & \bf 32.6\%
&  \bf 48.74& \bf 45.0\%\\
\midrule

\multirow{4}{*}{Llama-3.2-3B-Instruct}
& Before Distill.
& 44.46 & 13.8\%
& 45.72 & 8.0\%
& 47.25 & 16.5\%
& 47.98 & 5.0\% \\
& SeqKD
& 46.68 & 23.6\%
& 46.28 & 9.4\%
& 47.35 & 17.8\% 
& 48.49 & 18.8\% \\
& GAD
& 47.74 & 45.3\%
& 47.17 & 46.0\%
& 48.69 & 49.6\%
& 49.89 & 58.8\% \\
& \textbf{\ours}
& \bf48.97& \bf56.4\%
& \bf49.57 & \bf 64.6\%
& \bf50.01 & \bf62.4\% 
& \bf51.49& \bf78.8\%\\
\bottomrule
\end{tabular}
}
\label{tab:auto2}
\end{table}

\section{Experiments}
\subsection{Experimental Settings}
\textbf{Datasets.} 
We evaluate our method on two complementary datasets for assistant response generation, covering distinct data distributions.
(1) \textbf{LMSYS-Chat}: We primarily use a subset of the LMSYS-Chat-1M dataset~\cite{lmsys} preprocessed by GAD~\cite{gad}, which represents large-scale, diverse, and spontaneous user-chatbot interactions.
(2) \textbf{Dolly-Train}: To further evaluate performance on human-authored instruction data, we adopt the MiniLLM-processed Dolly training data released on Hugging Face~\cite{gu2024minillm}.

\textbf{Models.} 
We evaluate our method under black-box distillation from two API-based teachers, \textbf{GPT-5}~\cite{gpt5} and \textbf{Doubao-Seed-2.0}~\cite{Seed2}. Teacher responses are generated through their official APIs, with thinking mode enabled for Doubao-Seed-2.0. As students, we use two small-model families: \textbf{Qwen2.5-Instruct} at 1.5B and 3B scales~\cite{qwen}, and \textbf{Llama-3.2-Instruct} at 1B and 3B scales~\cite{grattafiori2024llama}.

\textbf{Baselines and Evaluation.} 
We compare GRGC with offline-teacher-response black-box distillation baselines, i.e., methods that learn from pre-generated teacher outputs without additional online teacher queries: SeqKD~\cite{seqkd} and GAD~\cite{gad}.
Following GAD, evaluation covers both in-distribution and OOD benchmarks: a 479-sample LMSYS held-out test set, DollyEval~\cite{dolly} with 500 samples, SelfInst~\cite{self_inst} with 242 user-oriented instructions, and VicunaEval~\cite{vicuna} with 80 complex open-ended queries.
We use Qwen2.5-72B~\cite{qwen} as the main LLM-as-a-judge, with details in Appendix~\ref{app:eval_detail}. Additional GPT-OSS-120B-based~\cite{agarwal2025gpt} and human evaluations are reported in Appendix~\ref{a4}.

\textbf{Implementation Details.} 
All main results are trained for a total of 2 epochs. For GAD and our approach, the process consists of 1 epoch of supervised warmup followed by 1 epoch of adversarial training. We use a global batch size of 128. Following GAD~\cite{gad}, the learning rate is set to $5 \times 10^{-6}$ for SeqKD, and $1 \times 10^{-6}$ for both stages in GAD and our method. The group size is $N=8$, the KL weight is $\beta=0.001$, and the training temperature is set to 0.8. The critic is initialized from the student model’s parameters and extended with an additional head to serve as the reward model. The maximum context length is 2,048 tokens for prompts and 1,536 tokens for responses, with responses from the Doubao‑Seed‑2.0 teacher capped at 2,048 tokens. Our OT weight is $\lambda_{\text{OT}} = 0.01$, and the group power hyperparameter is $\gamma = 1.5$. Experiments are conducted on 8 NVIDIA A100 GPUs.

\begin{table}[t]
\caption{Automatic evaluation results for models distilled from Doubao-Seed-2.0 and trained on Dolly Train. We report the averaged Qwen2.5-72B evaluation score and win rate on the test datasets.}
\centering
\resizebox{\linewidth}{!}{
\begin{tabular}{cc|cc|cc|cc|cc}
\toprule
\multirow{2}{*}{Model} & \multirow{2}{*}{Method}
& \multicolumn{2}{c|}{LMSYS}
& \multicolumn{2}{c|}{Dolly}
& \multicolumn{2}{c|}{SelfInst}
& \multicolumn{2}{c}{Vicuna} \\
&
& Score & Win
& Score & Win
& Score & Win
& Score & Win \\
\midrule
Doubao-Seed-2.0 & Teacher
& 53.23 & 82.5\%
& 53.14 & 75.8\%
& 52.70 & 74.0\%
& 53.20 & 93.8\%  \\
\midrule

\multirow{4}{*}{Qwen2.5-3B-Instruct}
& Before Distill.
& 45.79 & 11.9\%
& 44.93 & 4.0\%
& 46.56 & 12.8\%
& 47.85 & 3.8\% \\
& SeqKD
& 46.65 & 31.9\%
& 46.60 & 33.2\%
& 46.63 & 28.9\%
& 48.36 & 36.3\% \\
& GAD
& 47.92 & 33.6\%
& 46.91 & 35.2\%
& 48.38 & 36.4\%
& 50.16 & 55.0\% \\
& \textbf{\ours}
& \bf 48.93 & \bf45.1\%
& \bf 47.88 & \bf 42.8\%
& \bf 49.14 & \bf 43.4\%
&  \bf 51.25 & \bf 62.5\%  \\
\midrule

\multirow{4}{*}{Qwen2.5-1.5B-Instruct}
& Before Distill.
& 41.93 & 4.8\%
& 39.78 & 0.6\%
& 41.06 & 5.0\%
& 43.24 & 0.0\% \\
& SeqKD
& 42.19 & 14.2\%
& 43.30 & 19.6\%
& 43.70 & 21.9\%
& 45.50 & 15.0\% \\
& GAD
& 44.51 & 19.4\%
& 43.95 & 20.0\%
& 45.83 & 30.6\%
& 47.29 & 25.0\% \\
& \textbf{\ours}
& \bf 45.10 & \bf 22.1\% 
& \bf 44.82 & \bf 29.2\%
& \bf 46.73 & \bf 32.2\%
& \bf 48.30 & \bf 33.8\%\\
\midrule
\multirow{4}{*}{Llama-3.2-3B-Instruct}
& Before Distill.
& 44.46 & 13.8\%
& 45.72 & 8.0\%
& 47.25 & 16.5\%
& 47.98 & 5.0\% \\
& SeqKD
& 45.43 & 28.6\%
& 47.53 & 30.8\%
& 47.96 & 33.1\%
& 49.43 & 51.3\%\\
& GAD
& 45.47 & 30.3\%
& 49.04 & 41.4\%
& 48.70 & 39.3\%
& 50.15 & 56.3\% \\
& \textbf{\ours}
& \bf47.95 & \bf45.5\%
& \bf 49.71& \bf 57.4\%
& \bf 49.86& \bf 56.6\%
& \bf51.26 & \bf70.0\% \\
\bottomrule
\end{tabular}
}
\label{tab:auto_dolly}
\vspace{-0.5em}
\end{table}

\begin{table}[tb]
\centering

\begin{minipage}[t]{0.66\linewidth}
\centering
\small
\caption{Ablation study of models distilled from GPT-5-Chat and trained on the LMSYS-Chat across adversarial training epochs.}
\label{tab:epoch_auto_eval}
\resizebox{\linewidth}{!}{
\begin{tabular}{cc|cc|cc}
\toprule
\multirow{2}{*}{CGC} & \multirow{2}{*}{PGM} 
& \multicolumn{2}{c|}{\textbf{Adv. Epoch = 1}} 
& \multicolumn{2}{c}{\textbf{Adv. Epoch = 2}} \\
\cmidrule(lr){3-4} \cmidrule(lr){5-6}
& 
& Qwen2.5-3B & Qwen2.5-1.5B 
& Qwen2.5-3B & Qwen2.5-1.5B \\
\midrule
$\times$ & $\times$ & 48.44 & 46.05 & 48.52 & 46.37 \\
$\checkmark$ & $\times$ & 47.52 & 46.40 & 49.60 & 47.38 \\
$\times$ & $\checkmark$ & 48.88 & 46.44 & 48.95 & 46.69 \\
$\checkmark$ & $\checkmark$ & 50.19 & 47.36 & 50.27 & 47.70 \\
\bottomrule
\end{tabular}
}
\end{minipage}
\hfill
\begin{minipage}[t]{0.325\linewidth}
\centering
\small
\caption{Comparison with critic-side calibration alternatives.}
\label{tab:ablation_cgc}
\resizebox{\linewidth}{!}{
\begin{tabular}{c|cc}
\toprule
Distribution & LMSYS & Others \\
\midrule
Variance only & 46.83 & 46.61 \\
Linear & 47.00 & 46.79 \\
Laplace & 46.52 & 46.32 \\
Logistic & 46.54 & 46.45 \\
Gaussian & 47.36 & 47.13 \\
\bottomrule
\end{tabular}
}
\end{minipage}
\vspace{-0.5em}
\end{table}

\begin{table}[t]
\centering

\begin{minipage}[t]{0.36\linewidth}
\centering
\small
\caption{Ablation on the policy-side group modulation transformation.}
\vspace{0.2em}
\label{tab:ablation_PGM}
\resizebox{\linewidth}{!}{
\begin{tabular}{c|cc}
\toprule
Transformation & LMSYS & Others \\
\midrule
CDF & 46.63 & 46.74 \\
Top-1 & 46.44 & 46.41 \\
Rank & 46.69 & 46.35 \\
Group Power & 47.36 & 47.13 \\
\bottomrule
\end{tabular}
}
\end{minipage}
\hfill
\begin{minipage}[t]{0.61\linewidth}
\centering
\small
\caption{Average per-step training time, module-wise overhead, and combined overhead ratio across student model sizes.}
\vspace{0.2em}
\label{tab:time_overhead}
\resizebox{\linewidth}{!}{
\begin{tabular}{lcc}
\toprule
Metric & Qwen2.5-1.5B-Instruct & Qwen2.5-3B-Instruct \\
\midrule
Avg. step time & 37.54 s & 61.93 s \\
CGC overhead & 0.03 s & 0.03 s \\
PGM overhead & 0.01 s & 0.01 s \\
Overhead ratio & 0.11\% & 0.06\% \\
\bottomrule
\end{tabular}
}
\end{minipage}

\end{table}

\subsection{Results and Discussions}
\textbf{GPT-5-Teacher Results.}
Table~\ref{tab:auto1} shows that GRGC achieves the best 
results across all reported students and test sets.
This pattern appears in both averaged scores and reference-based win rates, indicating consistent gains across metrics and stronger preference relative to the reference answers.
These results are consistent with our central claim that better reward geometry improves grouped optimization and adversarial black-box distillation.
GPT-OSS-120B judging and human preference evaluation show the same trend as the 
Qwen2.5-72B evaluation.
Appendix~\ref{a4} reports these evaluations, along with judge-free IFEval, Math500, and long-form mathematical-reasoning tests, sensitivity analyses on $\lambda_{\text{OT}}$ and $\gamma$, and token-length analysis.
The confidence intervals in Tables~\ref{seedscore} and~\ref{seedwinrate} further demonstrate the robustness of our improvement.

\textbf{Doubao-Seed-2.0 Teacher Results.}
With thinking mode enabled, Doubao-Seed-2.0 produces longer and more structured responses, making it a more challenging teacher to distill.
Comparing the LMSYS-Chat results in Tables~\ref{tab:auto1} and~\ref{tab:auto2}, Doubao-Seed-2.0 achieves higher teacher win rates than GPT-5-Chat.
GRGC helps student models better approximate Doubao-Seed-2.0's response behavior, as reflected by improved reference-based win rates.
Table~\ref{tab:auto_dolly} shows that GRGC maintains its advantage when trained on Dolly-Train, demonstrating generalization beyond the LMSYS dataset distribution.

\textbf{Impact of the two components.}
Table~\ref{tab:epoch_auto_eval} verifies the intended interaction between CGC and PGM. CGC alone gives strong results after two adversarial epochs, showing that critic-side geometry calibration improves optimization by preventing reward collapse and scale instability. Its 1-epoch gain is smaller, and on Qwen2.5-3B it is even below GAD, because early OT calibration mainly organizes the reward geometry rather than immediately making the top student candidates more separated. PGM addresses exactly this early-stage gap: after group centering and scaling, the signed power transform preserves the critic-induced order while enlarging informative standardized differences, producing more decisive advantage gaps for the policy update. Combining the two therefore gives both well-conditioned critic rewards from CGC and stronger early advantage separation from PGM.

\begin{table}[t]
\centering
\begin{minipage}[t]{0.49\linewidth}
\centering
\captionof{table}{Common-normalized final-advantage geometry over all 192,000 groups in the complete Qwen2.5-3B/GPT-5/LMSYS trajectory.}
\label{tab:advantage_direct}
\small
\setlength{\tabcolsep}{1.5pt}
\begin{tabular*}{\linewidth}{@{\extracolsep{\fill}}lccc@{}}
\toprule
Metric & GAD & GRGC & Change \\
\midrule
Mean top-2 gap                    & 0.569  & \textbf{0.835}  & $+46.7\%$ \\
Median top-2 gap                  & 0.431  & \textbf{0.674}  & $+56.4\%$ \\
Weak ($\Delta_{1,2}\leq0.1$)     & 17.12\% & \textbf{13.23\%} & $-22.7\%$ \\
Top-1 (5\% noise)                & 87.44\% & \textbf{93.76\%} & $+6.32$ pp \\
Pairwise (5\% noise)             & 96.70\% & \textbf{98.36\%} & $+1.66$ pp \\
Top-1 (10\% noise)               & 79.54\% & \textbf{91.32\%} & $+11.78$ pp \\
\bottomrule
\end{tabular*}
\end{minipage}
\hfill
\begin{minipage}[t]{0.49\linewidth}
\centering
\captionof{table}{Effect of doubling the prompt-batch or response-group size for Llama-3.2-1B. Low dispersion denotes relative std $<0.15$.}
\label{tab:batch_group_scaling}
\small
\setlength{\tabcolsep}{1.5pt}
\begin{tabular*}{\linewidth}{@{\extracolsep{\fill}}llcc@{}}
\toprule
Setting & Method & Low-disp. & LMSYS \\
\midrule
Base & GAD & 7.98\% & 42.57 \\
Base & \textbf{GRGC} & \textbf{0.63\%} & \textbf{44.68} \\
$2\times$ group size & GAD & 6.83\% & 42.71 \\
$2\times$ group size & \textbf{GRGC} & \textbf{0.03\%} & \textbf{45.12} \\
$2\times$ batch size & GAD & 7.92\% & 42.31 \\
$2\times$ batch size & \textbf{GRGC} & \textbf{0.68\%} & \textbf{44.40} \\
\bottomrule
\end{tabular*}
\end{minipage}
\end{table}

\begin{wrapfigure}{r}{0.4\textwidth}
\centering
\vspace{-1em}
\includegraphics[width=\linewidth]{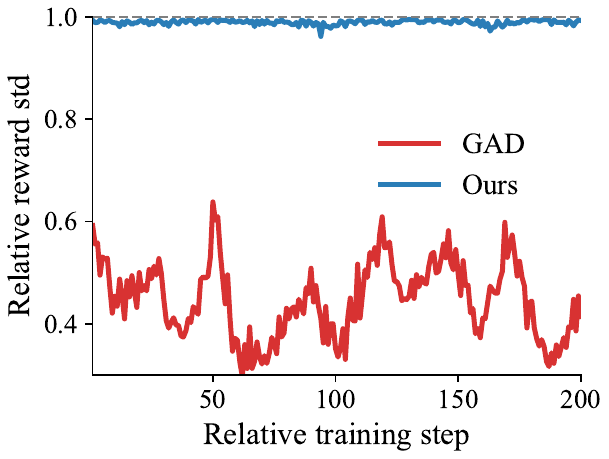}
\caption{Mean relative prompt-group reward spread, computed as the mean group reward standard deviation normalized by the global reward standard deviation at each step.
Our approach maintains a more stable reward scale.}
\label{fig:relstd_compare}
\vspace{-2em}
\end{wrapfigure}

\textbf{Discussion of CGC.}
Table~\ref{tab:ablation_cgc} shows that Gaussian CGC gives the strongest results among the tested conditioning priors. 
Variance-only calibration helps by reducing the most obvious collapse, but it does not constrain how gaps are allocated across ranks. Linear targets impose non-collapsed but overly rigid equal spacing, while Laplace and logistic targets introduce heavier tails that allocate too much separation to extreme-ranked samples. Gaussian OT gives the best trade-off by matching ordered rewards to a non-collapsed, symmetric, and smoothly varying reference geometry with moderate tail emphasis.
Details of these alternatives and more mechanism analysis are provided in Appendix~\ref{a2}.

\textbf{Discussion of PGM.}
Table~\ref{tab:ablation_PGM} shows that group power is the strongest reward-shaping choice among the tested alternatives. Its advantage is that it preserves the critic-implied ordering while placing larger gain on already separated grouped margins. This keeps the critic's magnitude information while avoiding the overly sparse update induced by Top-1 and the margin-information loss induced by CDF or Rank~\cite{choi2026gopo} remapping. More mechanism-level comparison is provided in Appendix~\ref{a2}. 

\textbf{Runtime analysis.}
GRGC operates on the scalar reward groups already produced by the critic, adding only sorting, normalization, OT matching, and elementwise reward modulation without introducing extra model forward or backward passes. Table~\ref{tab:time_overhead} confirms that this low-order overhead is negligible and largely insensitive to model size: the measured overhead is only $0.11\%$ for Qwen2.5-1.5B and $0.06\%$ for Qwen2.5-3B. A more detailed complexity analysis is provided in Appendix~\ref{a5}.

\textbf{Visualizations.}
The qualitative geometry diagnostics align closely with the quantitative gains. Fig.~\ref{fig:relstd_compare} shows that GRGC keeps the relative prompt-group reward spread within a more controlled range than GAD, suggesting improved critic-side conditioning for effective advantage construction. Figure~\ref{fig:ours_geometry_window} shows that GRGC reduces collapsed prompt groups, while Fig.~\ref{fig:weakmargin_baseline_vs_ours} shows that it also reduces weak-margin groups. 
Figure~\ref{fig:absolute_spread_compare} further reports the unnormalized prompt-group reward spread with uncertainty bands. 
These diagnostics support the proposed reward-geometry conditioning mechanism.

\textbf{Direct advantage analysis.}
Table~\ref{tab:advantage_direct} reconstructs both methods' final advantages using their respective reward transformations under the same population z-score convention. The top-2 gap is the difference between the largest and second-largest advantages within each prompt group. 
GRGC yields larger gaps, fewer near ties, and stronger top-1 and pairwise retention, demonstrating better separation and perturbation stability of the final GRPO advantages.
Table~\ref{tab:batch_group_scaling} further shows that additional sampling alone does not resolve low dispersion. Here batch size counts prompts per update, group size counts responses per prompt, and low dispersion denotes relative std $<0.15$. Increasing either the number of prompts per update or the number of responses per prompt provides only limited improvement for GAD, whereas GRGC consistently maintains substantially lower low-dispersion frequency and higher LMSYS performance across all sampling settings.

\section{Conclusion}
We identify a previously overlooked critic-to-advantage mismatch in adversarial black-box distillation: the Bradley--Terry critic optimizes teacher--student discrimination, whereas GRPO depends on the within-group geometry of student rewards. This diagnosis yields a two-stage reward-geometry principle, instantiated by GRGC through source-mean-anchored groupwise OT during critic-side reward formation and order-preserving contrast reshaping before final advantage normalization. Across model pairs, datasets, and evaluation protocols, GRGC produces better separated and more perturbation-stable advantages, improves distillation quality, and adds negligible runtime overhead.

\bibliography{reference_full}

\newpage
\appendix

\noindent\textbf{\Large Appendix}
\section{Overview}
This appendix provides supplementary materials that further elaborate on the related work, mechanism analysis, implementation details, evaluation protocol, and empirical findings of GRGC. It includes the following sections:

\begin{itemize}
    \item \textbf{Section~\ref{a1}: More Related Work.} Detailed discussions on prior studies, with an emphasis on optimal transport and related reward-shaping perspectives.

    \item \textbf{Section~\ref{a2}: Mechanism Analysis for OT Calibration and Group Power.} Formal analyses of the BT objective's fixed-mean collapse bias, the downstream effects of poor reward geometry, the scale, shape, and anchoring properties of Gaussian OT calibration, the signed power transform, and the effect of advantage errors on the clipped GRPO surrogate.

    \item \textbf{Section~\ref{a3}: Symbol Description.} A complete summary of key mathematical notations, hyperparameters, and definitions referenced throughout the paper.

    \item \textbf{Section~\ref{app:pseudocode}: Training Pseudocode.} Step-by-step pseudocode for the GRGC training pipeline, including the warmup phase, critic-side OT calibration, group-power reward shaping, and GRPO-based student updates.

    \item \textbf{Section~\ref{app:eval_detail}: Automatic Evaluation Details.} Full descriptions of the evaluation prompts, normalized score definition, and win-rate calculation used in automatic evaluation.

   \item \textbf{Section~\ref{a4}: Further Experimental Analyses.} Additional experiments and analyses beyond the main paper, including sensitivity analysis, experiments with step-by-step teacher prompting, human evaluation, visualization analyses, GPT-OSS evaluation, token-length analysis, rule-based IFEval, Math500 and long-form mathematical reasoning, and multi-seed evaluation robustness.

    \item \textbf{Section~\ref{a5}: Complexity Analysis.} A detailed comparison of the training complexity of GAD and GRGC, including the costs of BT loss, GRPO normalization, OT calibration, and group power shaping.

    \item \textbf{Section~\ref{a6}: Broader Impact.} Reflections on the broader societal, ethical, and practical implications of more effective black-box distillation.

    \item \textbf{Section~\ref{a7}: Limitations.} Critical discussion of the limitations of our framework.
\end{itemize}
Together, these supplementary materials provide a transparent view of the method, support reproducibility, and offer additional evidence that complements the main paper.
\section{More Related Work}
\label{a1}

\paragraph{Optimal Transport} Optimal Transport (OT) provides a principled geometric framework for comparing probability distributions by computing the minimal cost required to transform one distribution into another. Compared to divergences such as KL and Jensen--Shannon, OT remains meaningful even when supports do not overlap and thus yields a more faithful notion of distributional discrepancy~\cite{wd,tnnls3}. The resulting Wasserstein distance has been widely adopted in image generation~\cite{wgan1,wgan2,wgan3}, causal structure learning and reasoning~\cite{causal1,causal}, unsupervised learning~\cite{un1,un2,un4}, and reinforcement learning~\cite{re1,re2,re3}. Entropy-regularized OT, or Sinkhorn distance~\cite{sinkhornbase}, further enables scalable OT-based methods in domain adaptation~\cite{otda,zeng2024hierarchical,da2} and dataset distillation~\cite{cui2026optimal,cui2026optimizing,cui2025optical,cui2026geometry}. 

In knowledge distillation, OT has been used to transfer richer structural information than pointwise KL matching~\cite{sink,sink2}. WCoRD aligns teacher and student representation distributions with Wasserstein contrastive objectives~\cite{chen2021wcord}; KNOT distills teacher label distributions for NLP tasks by minimizing OT cost in label space~\cite{bhardwaj2022knot}; and recent LLM distillation methods use Wasserstein, OT, or cross-tokenizer distribution-matching objectives to handle vocabulary mismatch and multi-level token/sequence alignment~\cite{boizard2024towards,cui2025multilevel,cui2026flexible,minixhofer2025universal,vuong2026mcw}. These works use distribution matching as a teacher--student knowledge-transfer loss, typically aligning logits, label distributions, hidden representations, or token spaces. Our use of OT is different: GRGC does not align student outputs to teacher logits or representations. Instead, it applies a one-dimensional groupwise OT regularizer to the adversarial critic's student-side reward group, so that the reward geometry consumed by GRPO has controlled spread, ordered shape, and stable pre-normalization conditioning.

\paragraph{Reward Shaping and Reward Calibration} Reward shaping has long been used to improve reinforcement-learning optimization by modifying the reward signal while preserving or clarifying the target behavior. Potential-based reward shaping gives policy-invariance guarantees for MDPs~\cite{ng1999policy}, with later extensions connecting shaping to value initialization~\cite{wiewiora2003potential} and dynamic shaping functions~\cite{devlin2012dynamic}. Other classical methods address reward scale and credit assignment: PopArt normalizes value targets across changing reward magnitudes~\cite{vanhasselt2016popart}, while RUDDER redistributes delayed returns to reduce temporal credit-assignment difficulty~\cite{arjona2019rudder}. In language-model alignment, preference-based reward learning trains scalar reward models from comparisons~\cite{christiano2017deep,ziegler2019finetuning,ouyang2022training}, with later work studying reward overoptimization~\cite{gao2023scaling}, implicit reward modeling through DPO~\cite{rafailov2023direct}, and reward-model evaluation through RewardBench~\cite{lambert2024rewardbench}. Recent calibration-oriented work further shows that reward-model accuracy and reward-model confidence biases can be misaligned with downstream policy quality~\cite{chen2024accuracy,leng2025taming}, proposes post-hoc correction of reward-model biases such as length bias~\cite{huang2025posthoc}, and introduces OT-based distributional preference alignment~\cite{melnyk2024distributional}. These methods mainly study environment-reward transformations, value-target normalization, delayed-return redistribution, fixed/implicit reward models, or distributional alignment between preferred and dispreferred samples. GRGC addresses a different interface: in adversarial black-box distillation, the critic is trained online as a teacher--student discriminator and its student-side scores are immediately consumed by grouped advantage construction. Our calibration therefore targets the prompt-wise geometry of student-only reward groups, including their spread, ordered shape, and margin separability, rather than global reward scale, reward-model accuracy, or preference-distribution dominance alone.

\section{Mechanism Analysis for OT Calibration and Group Power}
\label{a2}
Here we formalize the mechanism behind GRGC across critic training, reward geometry, advantage construction, and grouped policy optimization. To keep the notation aligned with the main method, we distinguish four quantities throughout this section: raw critic rewards $r_i$, intermediate standardized grouped scores $z_i$, transformed rewards $\hat r_i$, and final GRPO advantages $A_i$. The mechanism we analyze is therefore
\begin{equation}
\begin{aligned}
&\text{BT critic objective}
\;\longrightarrow\;
\text{reward geometry of } r
\;\longrightarrow\;
\text{stability of } z
\;\longrightarrow\;
\text{separability of } \hat r \\
&\;\longrightarrow\;
\text{quality of the final advantages } A
\;\longrightarrow\;
\text{policy separation and clipped-surrogate stability}.
\end{aligned}
\end{equation}
This section establishes the complete mechanism chain. We first show that the BT critic objective has an intrinsic fixed-mean bias toward collapsed student reward groups, and then quantify how low dispersion and weak scale make grouped signals fragile. We next show that Gaussian OT calibration controls scale and ordered shape while using the unique optimal translation anchor, after which the signed power transform enlarges already informative margins. Finally, we connect the resulting advantage gaps to one-step policy separation and bound the effect of advantage errors on the clipped GRPO surrogate.

\paragraph{Setup.}
Consider a prompt group with raw critic rewards $\mathcal{G}(x)=\{r_i\}_{i=1}^N$ and group mean
\begin{equation}
\mu_x=\frac{1}{N}\sum_{i=1}^N r_i.
\end{equation}
Let the ordered centered rewards be
\begin{equation}
u_{(j)}=r_{(j)}-\mu_x,\qquad j=1,\dots,N,
\end{equation}
and denote their empirical standard deviation by
\begin{equation}
\sigma_x^2=\frac{1}{N}\sum_{j=1}^N u_{(j)}^2.
\end{equation}
The intermediate standardized grouped scores used by the power transform are
\begin{equation}
z_i(r)=\frac{r_i-\mu_x}{\sigma_x+\varepsilon}.
\end{equation}
The transformed rewards are then
\begin{equation}
\hat r_i=f_\gamma(z_i)=\mathrm{sign}(z_i)|z_i|^\gamma,
\end{equation}
and the final GRPO advantages constructed from $\hat r$ are
\begin{equation}
A_i(\hat r)=\frac{\hat r_i-\frac{1}{N}\sum_{j=1}^N \hat r_j}{\sqrt{\frac{1}{N}\sum_{j=1}^N \left(\hat r_j-\frac{1}{N}\sum_{\ell=1}^N \hat r_\ell\right)^2}+\varepsilon}.
\end{equation}
This makes three aspects of reward geometry immediately relevant:
\begin{itemize}
\item \emph{ranking stability}: whether the within-group ordering is robust to perturbations;
\item \emph{signal sensitivity}: how strongly perturbations in raw rewards are transmitted into the constructed grouped signal.
\item \emph{margin separability}: whether informative candidates remain sufficiently separated after transformation.
\end{itemize}

\paragraph{Proposition 1: the BT objective has a fixed-mean group-collapse bias.}
Fix a prompt $x$, its teacher critic score $D(y)$, and a student group mean $\mu_x$. For any student score vector $r=(r_1,\ldots,r_N)$ satisfying $\frac{1}{N}\sum_i r_i=\mu_x$, define its prompt-wise BT loss by
\begin{equation}
\mathcal{B}_y(r)
=
\frac{1}{N}\sum_{i=1}^N \log\!\left(1+\exp(r_i-D(y))\right).
\end{equation}
Then
\begin{equation}
\mathcal{B}_y(r)
\ge
\log\!\left(1+\exp(\mu_x-D(y))\right),
\end{equation}
with equality if and only if
\begin{equation}
r_1=\cdots=r_N=\mu_x.
\end{equation}
Thus, for fixed teacher score and fixed student group mean, the BT objective uniquely minimizes its loss at zero within-group dispersion.

\paragraph{Proof.}
Let $\phi_y(s)=\log(1+\exp(s-D(y)))$. Its second derivative is
\begin{equation}
\phi_y''(s)=\frac{\exp(s-D(y))}{(1+\exp(s-D(y)))^2}>0,
\end{equation}
so $\phi_y$ is strictly convex. Jensen's inequality gives
\begin{equation}
\frac{1}{N}\sum_{i=1}^N\phi_y(r_i)
\ge
\phi_y\!\left(\frac{1}{N}\sum_{i=1}^N r_i\right)
=
\phi_y(\mu_x).
\end{equation}
Strict convexity makes equality possible only when all $r_i$ are identical, and the mean constraint then forces their common value to be $\mu_x$.

\paragraph{Interpretation.}
The mismatch is stronger than a missing regularizer: on every fixed-mean slice of student scores, the BT term strictly penalizes mean-preserving dispersion and uniquely favors collapse. This does not claim that unconstrained adversarial training must always collapse, because the teacher score and group mean also evolve. It establishes that the critic objective contains no countervailing within-group preference and instead exerts a structural pressure toward the failure mode diagnosed in Figure~\ref{fig:gad_pathology}.

\paragraph{Proposition 2: low dispersion makes grouped ranking fragile.}
Assume perturbed rewards $\tilde r_i=r_i+\eta_i$ with $|\eta_i|\le \eta$ for all $i$. Let
\begin{equation}
\Delta_{\min}(r)=\min_{i\neq j,\;r_i\neq r_j}|r_i-r_j|
\end{equation}
be the smallest nonzero within-group margin, with the convention $\Delta_{\min}(r)=0$ if all rewards are tied. If
\begin{equation}
\Delta_{\min}(r)>2\eta,
\end{equation}
then every originally strict pairwise comparison is preserved under perturbation. Exact ties are not covered by this guarantee and may be broken by perturbations. Equivalently, any change to an originally strict comparison must satisfy
\begin{equation}
\Delta_{\min}(r)\le 2\eta.
\end{equation}

\paragraph{Proof sketch.}
For any pair $i,j$ with $r_i>r_j$,
\begin{equation}
\tilde r_i-\tilde r_j=(r_i-r_j)+(\eta_i-\eta_j)\ge (r_i-r_j)-2\eta.
\end{equation}
Hence if every nonzero pairwise gap exceeds $2\eta$, the sign of every originally nonzero pairwise difference is preserved. Conversely, changing any originally strict pairwise order is possible only if at least one original nonzero gap is at most $2\eta$.

\paragraph{Interpretation.}
Proposition 2 formalizes the most direct downstream consequence of low dispersion and weak margins. When a reward group becomes too flat, the smallest useful within-group gap becomes comparable to ordinary critic noise. The result is not merely a cosmetic change in summary statistics: the identity of the better student response can itself become unstable.

\paragraph{Proposition 3: low scale makes the standardized grouped signal more sensitive.}
Let $z(r)=(z_1(r),\dots,z_N(r))^\top$ and $\tilde z=z(\tilde r)$ with $\tilde r=r+\zeta$. Write $\tilde\mu_x=\frac{1}{N}\sum_i \tilde r_i$ and $\tilde\sigma_x^2=\frac{1}{N}\sum_i (\tilde r_i-\tilde\mu_x)^2$ for the perturbed group mean and perturbed group scale. Assume the perturbation is local in the sense that
\begin{equation}
\|\zeta\|_2 \le \frac{\sqrt{N}}{2}(\sigma_x+\varepsilon).
\end{equation}
Then
\begin{equation}
\|\tilde z-z\|_2
\le
\frac{2\|\zeta\|_2}{\sigma_x+\varepsilon}
\left(
1+\frac{\sigma_x}{\sigma_x+\varepsilon}
\right)
\le
\frac{4\|\zeta\|_2}{\sigma_x+\varepsilon}.
\end{equation}
In particular, for fixed perturbation magnitude $\|\zeta\|_2$, the perturbation in the standardized grouped signal grows as the pre-normalization group scale $\sigma_x$ becomes small.

\paragraph{Proof sketch.}
Write
\begin{equation}
z(r)=\frac{P r}{\sigma_x+\varepsilon},
\qquad
P=I-\frac{1}{N}\mathbf{1}\mathbf{1}^\top.
\end{equation}
Then
\begin{equation}
\tilde z-z
=
\frac{P(r+\zeta)}{\tilde\sigma_x+\varepsilon}
-\frac{Pr}{\sigma_x+\varepsilon}.
\end{equation}
Add and subtract $\frac{Pr}{\tilde\sigma_x+\varepsilon}$ to obtain
\begin{equation}
\tilde z-z
=
\frac{P\zeta}{\tilde\sigma_x+\varepsilon}
+
Pr\left(
\frac{1}{\tilde\sigma_x+\varepsilon}
-\frac{1}{\sigma_x+\varepsilon}
\right).
\end{equation}
The empirical standard deviation is $1/\sqrt{N}$-Lipschitz with respect to the Euclidean norm, so
\begin{equation}
|\tilde\sigma_x-\sigma_x|\le \frac{\|\zeta\|_2}{\sqrt{N}}.
\end{equation}
Under the local perturbation assumption, this implies
\begin{equation}
\tilde\sigma_x+\varepsilon \ge \frac{1}{2}(\sigma_x+\varepsilon).
\end{equation}
Using $\|P\zeta\|_2\le \|\zeta\|_2$, $\|Pr\|_2=\|r-\mu_x\mathbf{1}\|_2$, and
\begin{equation}
\left|
\frac{1}{\tilde\sigma_x+\varepsilon}
-\frac{1}{\sigma_x+\varepsilon}
\right|
\le
\frac{|\tilde\sigma_x-\sigma_x|}{(\sigma_x+\varepsilon)(\tilde\sigma_x+\varepsilon)},
\end{equation}
we obtain
\begin{equation}
\left|
\frac{1}{\tilde\sigma_x+\varepsilon}
-\frac{1}{\sigma_x+\varepsilon}
\right|
\le
\frac{2\|\zeta\|_2}{\sqrt{N}(\sigma_x+\varepsilon)^2}.
\end{equation}
Therefore
\begin{equation}
\|\tilde z-z\|_2
\le
\frac{2\|\zeta\|_2}{\sigma_x+\varepsilon}
+
\frac{2\|r-\mu_x\mathbf{1}\|_2\,\|\zeta\|_2}{\sqrt{N}(\sigma_x+\varepsilon)^2},
\end{equation}
and the identity $\|r-\mu_x\mathbf{1}\|_2=\sqrt{N}\sigma_x$ gives
\begin{equation}
\|\tilde z-z\|_2
\le
\frac{2\|\zeta\|_2}{\sigma_x+\varepsilon}
\left(
1+\frac{\sigma_x}{\sigma_x+\varepsilon}
\right)
\le
\frac{4\|\zeta\|_2}{\sigma_x+\varepsilon}.
\end{equation}

\paragraph{Interpretation.}
Proposition 3 formalizes why unstable or collapsed reward scale directly hurts grouped signal construction. The issue is not simply that some groups have lower variance than others. Rather, once the pre-normalization scale becomes too small, the same raw perturbation produces a much larger change in the standardized grouped signal $z$, which is exactly the quantity later consumed by the power transform.

\paragraph{Proposition 4.1: OT calibration lower-bounds scale mismatch.}
Define the centered Gaussian target quantiles
\begin{equation}
q_j=\Phi^{-1}\!\left(\frac{j-0.5}{N}\right),
\qquad
t_j=\mu_x+q_j,
\end{equation}
and let $\nu_x^{(N)}=\frac{1}{N}\sum_{j=1}^N\delta_{t_j}$ be the discrete Gaussian-quantile reference used in training. The critic-side OT loss is the exact one-dimensional Wasserstein distance from the empirical reward distribution to this discrete reference:
\begin{equation}
\mathcal{L}_{\mathrm{OT}}(x)
=
W_2^2(\hat{\mu}_x,\nu_x^{(N)})
=
\frac{1}{N}\sum_{j=1}^N (r_{(j)}-t_j)^2
=
\frac{1}{N}\sum_{j=1}^N (u_{(j)}-q_j)^2.
\end{equation}
Let
\begin{equation}
\sigma_q^2=\frac{1}{N}\sum_{j=1}^N q_j^2.
\end{equation}
Assume $N\ge 2$, so $\sigma_q>0$. Then
\begin{equation}
\mathcal{L}_{\mathrm{OT}}(x)\ge (\sigma_x-\sigma_q)^2.
\end{equation}
In particular, if the reward group collapses to a constant, so that $\sigma_x=0$, then
\begin{equation}
\mathcal{L}_{\mathrm{OT}}(x)\ge \sigma_q^2>0.
\end{equation}

\paragraph{Proof sketch.}
Expanding the centered form gives
\begin{equation}
\mathcal{L}_{\mathrm{OT}}(x)
=
\sigma_x^2+\sigma_q^2-2\langle u,q\rangle_N,
\qquad
\langle u,q\rangle_N=\frac{1}{N}\sum_{j=1}^N u_{(j)}q_j.
\end{equation}
By Cauchy--Schwarz,
\begin{equation}
\langle u,q\rangle_N\le \sigma_x\sigma_q,
\end{equation}
which yields the stated lower bound.

\paragraph{Interpretation.}
Proposition 4.1 shows that CGC does not merely encourage a larger average variance. It penalizes collapse against a full ordered target geometry with strictly positive spread. This matters because Proposition 3 depends on the actual scale of the reward group consumed by normalization. OT calibration therefore improves the pre-normalization geometry that determines the stability of the downstream grouped signal.

\paragraph{Proposition 4.2: OT calibration also penalizes ordered shape mismatch.}
Assume $\sigma_x>0$ and $\sigma_q>0$. Define the normalized alignment between the ordered centered reward group and the centered Gaussian target quantiles by
\begin{equation}
\rho_x=
\left\langle
\frac{u}{\sigma_x},
\frac{q}{\sigma_q}
\right\rangle_N
=
\frac{1}{N}\sum_{j=1}^N
\frac{u_{(j)}}{\sigma_x}
\frac{q_j}{\sigma_q},
\end{equation}
Then
\begin{equation}
\mathcal{L}_{\mathrm{OT}}(x)
=
(\sigma_x-\sigma_q)^2
+
2\sigma_x\sigma_q(1-\rho_x).
\end{equation}
In particular, even when $\sigma_x=\sigma_q$,
\begin{equation}
\mathcal{L}_{\mathrm{OT}}(x)=2\sigma_x^2(1-\rho_x),
\end{equation}
so the OT loss still penalizes deviations in ordered within-group geometry unless the reward group is perfectly aligned with the target quantile shape.

\paragraph{Proof sketch.}
From the expansion
\begin{equation}
\mathcal{L}_{\mathrm{OT}}(x)=\sigma_x^2+\sigma_q^2-2\langle u,q\rangle_N,
\end{equation}
we write
\begin{equation}
\langle u,q\rangle_N
=
\sigma_x\sigma_q
\left\langle
\frac{u}{\sigma_x},
\frac{q}{\sigma_q}
\right\rangle_N
=
\sigma_x\sigma_q\rho_x.
\end{equation}
Substituting this identity gives the stated decomposition.

\paragraph{Interpretation.}
Proposition 4.2 directly answers why CGC is more than variance control. The first term measures scale mismatch, but the second term measures ordered shape mismatch: how the empirical within-group gaps align with the target quantile progression across ranks. A variance-only regularizer can constrain the first term, but it leaves the second completely unconstrained. OT calibration therefore regularizes the full one-dimensional prompt-group geometry rather than only its overall spread.

\paragraph{Proposition 4.3: group-mean anchoring is OT-optimal and translation invariant.}
Let $q_1\le\cdots\le q_N$ be any centered template with $\frac{1}{N}\sum_j q_j=0$; the Gaussian quantiles used by CGC satisfy this condition. Among all translations of this template, define
\begin{equation}
\mathcal{C}_x(a)
=
\frac{1}{N}\sum_{j=1}^N
\left(r_{(j)}-(a+q_j)\right)^2,
\qquad a\in\mathbb{R}.
\end{equation}
Then
\begin{equation}
\mathcal{C}_x(a)
=
\mathcal{C}_x(\mu_x)+(a-\mu_x)^2.
\end{equation}
Consequently, $a^*=\mu_x$ is the unique minimizer. Moreover, writing $\mathcal{L}_{\mathrm{OT}}(r)=\mathcal{C}_x(\mu_x)$, for every common shift $c\in\mathbb{R}$,
\begin{equation}
\mathcal{L}_{\mathrm{OT}}(r+c\mathbf{1})
=
\mathcal{L}_{\mathrm{OT}}(r).
\end{equation}

\paragraph{Proof.}
Using $u_{(j)}=r_{(j)}-\mu_x$ and the centeredness of both $u$ and $q$,
\begin{align}
\mathcal{C}_x(a)
&=
\frac{1}{N}\sum_{j=1}^N
\left(u_{(j)}-q_j+\mu_x-a\right)^2 \\
&=
\frac{1}{N}\sum_{j=1}^N
\left(u_{(j)}-q_j\right)^2
+(\mu_x-a)^2,
\end{align}
because $\frac{1}{N}\sum_j(u_{(j)}-q_j)=0$. The decomposition and uniqueness follow immediately. For the invariance result, adding $c$ shifts every order statistic and the group mean by $c$, so
\begin{equation}
\left(r_{(j)}+c\right)-\left(\mu_x+c+q_j\right)
=
r_{(j)}-(\mu_x+q_j)
\end{equation}
for every $j$, leaving the OT loss unchanged.

\paragraph{Interpretation.}
Centering the reference at the current student group mean is therefore the unique least-cost translation of any prescribed centered geometry, rather than a heuristic anchor. It also makes CGC insensitive to an arbitrary common offset in critic scores. If the teacher and student critic scores are shifted together, the BT score differences are unchanged and Proposition~4.3 shows that the CGC term is unchanged as well; the combined critic objective can therefore regulate within-group geometry without imposing an absolute score origin.

\paragraph{Why Gaussian is a canonical reference rather than an arbitrary one.}
The Gaussian target is best understood as a canonical member of a broader ordered location-scale family. More generally, one could use any ordered location-scale family
\begin{equation}
\tilde t_j=\mu_x+s\,\psi_j,\qquad \psi_1\le\cdots\le\psi_N.
\end{equation}
The relevant question is which family best satisfies the structural requirements imposed by grouped advantage construction. We require four properties:
\begin{itemize}
\item non-degenerate spread, so collapse is penalized;
\item symmetry, so neither tail is privileged a priori;
\item smooth rank progression, so adjacent target gaps change gradually rather than abruptly;
\item single-parameter scale control, so the overall conditioning strength is easy to tune.
\end{itemize}
These desiderata are necessary but not sufficient to identify a unique target: other symmetric location-scale families, including logistic quantiles, also provide closed-form, non-degenerate, single-scale templates. The key distinction is therefore the induced rank-gap profile. We use the Gaussian target because it provides a moderate interior-to-tail progression: less rigid than constant linear spacing, but less tail-aggressive than heavier-tailed alternatives such as Laplace or logistic. In this sense, the Gaussian target is used as a canonical conditioning prior for grouped optimization geometry, not as a claim about the true semantic distribution of teacher preference.

\paragraph{Proposition 5: signed power preserves ordering and enlarges already informative gaps.}
Let
\begin{equation}
z_i=z_i(r)=\frac{r_i-\mu_x}{\sigma_x+\varepsilon},
\qquad
f_\gamma(z)=\mathrm{sign}(z)|z|^\gamma,\qquad \gamma>1.
\end{equation}
Then $f_\gamma$ is strictly monotone on $\mathbb{R}$ and therefore preserves within-group ordering. Moreover, if $z_i$ and $z_j$ have the same sign and satisfy
\begin{equation}
\min(|z_i|,|z_j|)\ge \rho,
\end{equation}
for some $\rho>0$, then
\begin{equation}
|f_\gamma(z_i)-f_\gamma(z_j)|
\ge
\gamma \rho^{\gamma-1}|z_i-z_j|.
\end{equation}

\paragraph{Proof sketch.}
For $z\neq 0$,
\begin{equation}
f_\gamma'(z)=\gamma |z|^{\gamma-1}>0,
\end{equation}
so the map is strictly monotone. If $z_i$ and $z_j$ share the same sign, the mean value theorem gives
\begin{equation}
|f_\gamma(z_i)-f_\gamma(z_j)|=f_\gamma'(\xi)|z_i-z_j|
\end{equation}
for some $\xi$ between $z_i$ and $z_j$. The lower bound follows because $|\xi|\ge \rho$ on that interval.

\paragraph{Interpretation.}
Proposition 5 moves the power-transform analysis beyond simple monotonicity. Once critic-side calibration has produced a group whose standardized scores have already exited the near-tie regime, group power increases the usable separation of those scores for advantage construction. This is exactly the quantity grouped optimization needs: not arbitrary rescaling, but stronger separation among already informative candidates.

\paragraph{Proposition 6: bounded-noise margin preservation in the informative regime.}
Assume latent standardized scores $z_i^\star$ and observed scores $z_i=z_i^\star+\xi_i$ with $|\xi_i|\le \eta$. Consider two candidates $i,j$ such that
\begin{equation}
z_i^\star>z_j^\star,\qquad z_j^\star\ge 1+\eta,
\end{equation}
and define the latent margin
\begin{equation}
\Delta_{ij}^\star=z_i^\star-z_j^\star.
\end{equation}
If
\begin{equation}
\Delta_{ij}^\star>2\eta,
\end{equation}
then the observed ordering is preserved and the transformed margin satisfies
\begin{equation}
|f_\gamma(z_i)-f_\gamma(z_j)|
\ge
\gamma(\Delta_{ij}^\star-2\eta).
\end{equation}

\paragraph{Proof sketch.}
As in Proposition 2,
\begin{equation}
z_i-z_j\ge \Delta_{ij}^\star-2\eta>0,
\end{equation}
so the observed ordering is unchanged. The condition $z_j^\star\ge 1+\eta$ implies $z_j\ge 1$ and therefore $z_i\ge z_j\ge 1$, so the interval between them lies entirely in a regime where $f_\gamma'(z)\ge \gamma$. Applying Proposition 5 gives the bound. The corresponding lower-tail case follows by applying the same argument to $-z_i$ and using the odd symmetry of $f_\gamma$.

\paragraph{Interpretation.}
This proposition clarifies the role of PGM. Once CGC has moved a prompt group into a regime where standardized margins are already informative, PGM preserves those rankings and enlarges the usable separation further. This makes the transformed reward signal $\hat r$ more decisive for final advantage construction; the resulting optimization benefit is then supported by the grouped-surrogate analysis below and the main experiments.

\paragraph{Proposition 7: advantage gaps control one-step policy separation in a grouped softmax surrogate.}
Consider a grouped softmax policy
\begin{equation}
\pi_i(\theta)=\frac{e^{\theta_i}}{\sum_{k=1}^N e^{\theta_k}},
\end{equation}
and a one-step grouped update driven by constructed advantages,
\begin{equation}
\theta_i^+=\theta_i+\eta_{\mathrm{pg}} A_i,
\qquad \eta_{\mathrm{pg}}>0.
\end{equation}
Then for every pair $i,j$,
\begin{equation}
\log\frac{\pi_i(\theta^+)}{\pi_j(\theta^+)}
-
\log\frac{\pi_i(\theta)}{\pi_j(\theta)}
=
\eta_{\mathrm{pg}}(A_i-A_j).
\end{equation}

\paragraph{Proof sketch.}
For the softmax parameterization,
\begin{equation}
\log\frac{\pi_i(\theta)}{\pi_j(\theta)}=\theta_i-\theta_j.
\end{equation}
Applying the update gives
\begin{equation}
\log\frac{\pi_i(\theta^+)}{\pi_j(\theta^+)}
=
(\theta_i+\eta_{\mathrm{pg}} A_i)-(\theta_j+\eta_{\mathrm{pg}} A_j),
\end{equation}
and subtracting the original log-odds yields the claim.

\paragraph{Interpretation.}
Proposition 7 links final advantage quality to a policy-level quantity in the grouped softmax surrogate. Pairwise policy separation after one grouped update is controlled by the corresponding pairwise advantage gap. Larger and more reliable final advantage gaps therefore produce larger and more reliable one-step separation toward the candidate assigned higher advantage in this surrogate model.

\paragraph{Proposition 8: ranking errors in the constructed advantages can induce misaligned one-step movement.}
Let $i^\star$ denote the best candidate under the underlying latent preference in a prompt group. Under the same one-step grouped update, for any competitor $j$,
\begin{equation}
\begin{aligned}
&\log\frac{\pi_{i^\star}(\theta^+)}{\pi_j(\theta^+)}
-
\log\frac{\pi_{i^\star}(\theta)}{\pi_j(\theta)} \\
&\qquad=
\eta_{\mathrm{pg}}(A_{i^\star}-A_j).
\end{aligned}
\end{equation}
Hence:
\begin{itemize}
\item if $A_{i^\star}>A_j$, the update increases the policy odds of the best candidate against $j$;
\item if $A_{i^\star}\le A_j$, the update fails to improve this pairwise preference and may move in the wrong direction.
\end{itemize}

\paragraph{Proof sketch.}
This is the special case of Proposition 7 obtained by setting $i=i^\star$. The sign of the one-step log-odds change is therefore exactly the sign of $A_{i^\star}-A_j$.

\paragraph{Interpretation.}
Proposition 8 links advantage quality to update direction in the grouped surrogate model. Once the constructed advantages become unstable or misranked, the grouped policy update itself can become misaligned. Combined with Propositions 1--7, this yields the mechanism-level chain
\begin{equation}
\begin{aligned}
\text{poor reward geometry}
&\;\Longrightarrow\;
\text{fragile or distorted advantages} \\
&\;\Longrightarrow\;
\text{weaker or wrong policy separation}.
\end{aligned}
\end{equation}
This is exactly the failure pattern visualized in Figure~\ref{fig:gad_pathology}.

\paragraph{Proposition 9: advantage errors bound clipped GRPO surrogate distortion.}
Consider a fixed prompt group sampled from the old policy $\pi_{\theta_{\mathrm{old}}}$, with positive old-policy probability for every sampled candidate.
Assume $0<\epsilon_{\mathrm{clip}}<1$.
For a candidate policy $\pi_\theta$, define the likelihood ratio
\begin{equation}
w_i(\theta)=
\frac{\pi_\theta(G(x)_i\mid x)}
{\pi_{\theta_{\mathrm{old}}}(G(x)_i\mid x)}
\end{equation}
and the clipped ratio
\begin{equation}
\bar w_i(\theta)=
\mathrm{clip}\!\left(w_i(\theta),1-\epsilon_{\mathrm{clip}},1+\epsilon_{\mathrm{clip}}\right).
\end{equation}
For an advantage vector $A\in\mathbb{R}^N$, the clipped grouped surrogate is
\begin{equation}
\mathcal{J}_{\mathrm{clip}}(\theta;A)
=
\frac{1}{N}\sum_{i=1}^N
\min\!\left(w_i(\theta)A_i,\bar w_i(\theta)A_i\right).
\end{equation}
Assume the local update region satisfies $0\le w_i(\theta)\le 1+\kappa$ for all $i$ and some $\kappa\ge \epsilon_{\mathrm{clip}}$.
Then for any two advantage vectors $A$ and $\tilde A$,
\begin{equation}
\left|
\mathcal{J}_{\mathrm{clip}}(\theta;A)
-
\mathcal{J}_{\mathrm{clip}}(\theta;\tilde A)
\right|
\le
\frac{1+\kappa}{N}\|A-\tilde A\|_1
\le
\frac{1+\kappa}{\sqrt{N}}\|A-\tilde A\|_2.
\end{equation}
Define the clean and distorted surrogate improvements by
\begin{equation}
\Delta_A(\theta)=
\mathcal{J}_{\mathrm{clip}}(\theta;A)
-
\mathcal{J}_{\mathrm{clip}}(\theta_{\mathrm{old}};A),
\qquad
\Delta_{\tilde A}(\theta)=
\mathcal{J}_{\mathrm{clip}}(\theta;\tilde A)
-
\mathcal{J}_{\mathrm{clip}}(\theta_{\mathrm{old}};\tilde A).
\end{equation}
Consequently, if $\Delta_A(\theta)\ge m$ and the constructed advantages satisfy $\|A-\tilde A\|_2\le \epsilon_A$, then $\Delta_{\tilde A}(\theta)>0$ whenever
\begin{equation}
m>
\frac{2(1+\kappa)}{\sqrt{N}}\epsilon_A.
\end{equation}
If both advantage vectors are group-centered, i.e., $\sum_i A_i=\sum_i\tilde A_i=0$, then the old-policy baseline terms vanish and the sufficient condition improves to
\begin{equation}
m>
\frac{1+\kappa}{\sqrt{N}}\epsilon_A.
\end{equation}

\paragraph{Proof sketch.}
For fixed $\theta$, both $w_i(\theta)$ and $\bar w_i(\theta)$ are nonnegative constants with
\begin{equation}
0\le w_i(\theta)\le 1+\kappa,
\qquad
0\le \bar w_i(\theta)\le 1+\epsilon_{\mathrm{clip}}\le 1+\kappa.
\end{equation}
For any nonnegative constants $a,b\le 1+\kappa$, the function
\begin{equation}
h(c)=\min(ac,bc)
\end{equation}
is $(1+\kappa)$-Lipschitz in $c$: for $c\ge 0$ it equals $\min(a,b)c$, and for $c<0$ it equals $\max(a,b)c$, so its slope always lies in $[0,1+\kappa]$.
Therefore,
\begin{equation}
\left|
\min(w_i A_i,\bar w_i A_i)
-
\min(w_i \tilde A_i,\bar w_i \tilde A_i)
\right|
\le
(1+\kappa)|A_i-\tilde A_i|.
\end{equation}
Averaging over $i$ gives the $\ell_1$ bound, and Cauchy--Schwarz gives the $\ell_2$ bound. For the preservation statement, apply this bound once to the candidate update and once to the old policy; the surrogate-improvement error is therefore at most $2(1+\kappa)\epsilon_A/\sqrt{N}$. If $A$ and $\tilde A$ are group-centered, then $w_i(\theta_{\mathrm{old}})=\bar w_i(\theta_{\mathrm{old}})=1$ for all $i$, so
\begin{equation}
\mathcal{J}_{\mathrm{clip}}(\theta_{\mathrm{old}};A)=\frac{1}{N}\sum_i A_i=0,
\qquad
\mathcal{J}_{\mathrm{clip}}(\theta_{\mathrm{old}};\tilde A)=\frac{1}{N}\sum_i \tilde A_i=0.
\end{equation}
Only the candidate-policy surrogate term needs to be controlled, giving the improved factor.

\paragraph{Interpretation.}
Proposition 9 gives an objective-level link between advantage quality and the actual clipped GRPO surrogate used for policy optimization. It shows that distorted advantages do not merely affect a ranking diagnostic: they perturb the clipped surrogate itself by an amount proportional to the advantage error. Because GRPO constructs group-centered advantages, the sharper preservation bound applies directly to our setting, and the same argument extends to minibatches by averaging over prompt groups. Thus, the preceding geometry results matter because CGC and PGM reduce the sources of advantage distortion that enter the real policy objective.

\paragraph{Why the two components are complementary.}
The two modules optimize successive links in the same mechanism. Proposition~1 identifies the BT objective's fixed-mean collapse bias, while Propositions~2 and 3 quantify the resulting ranking fragility and normalization sensitivity. CGC counteracts this source-side failure: Propositions~4.1--4.3 establish non-collapse, ordered-shape control, optimal anchoring, and common-shift invariance. PGM then acts on the conditioned grouped signal: Propositions~5--8 show that it preserves ordering, enlarges informative separation, and translates advantage gaps into correctly directed one-step policy movement. Proposition~9 further shows that advantage errors induce bounded distortion of the clipped GRPO surrogate. Together, these results formalize the full chain from critic-objective bias to reward geometry, advantage quality, and the policy objective; Figure~\ref{fig:gad_pathology} and the main ablations corroborate its optimization-level consequences.
\subsection{Why a Gaussian Target Among Alternative Conditioning Priors}
The OT construction in GRGC does not require the target geometry to be Gaussian in a literal semantic sense. More generally, one could choose any ordered reference quantiles
\begin{equation}
\tilde t_j=\mu_x+s\,\psi_j,\qquad j=1,\dots,N,
\end{equation}
where the template $\{\psi_j\}$ is centered, ordered, and non-degenerate. The question is therefore not whether other priors are possible, but why the Gaussian prior is a particularly suitable one for our setting.

\paragraph{Concrete target forms.}
To make the comparison explicit, let
\begin{equation}
p_j=\frac{j-0.5}{N},\qquad j=1,\dots,N.
\end{equation}
In our implementation, the alternative target families are written under a common nominal scale convention. For Gaussian, Laplace, and logistic targets this corresponds to unit population standard deviation, while the linear baseline is the bounded linearly spaced template used in the ablation. Therefore, the Linear variant should be read as an implemented bounded-spacing baseline, not as a unit-variance shape-matched prior. The concrete choices are
\begin{equation}
t_j^{\mathrm{lin}}=\mu_x+\ell_j,\qquad \ell_j\in\mathrm{linspace}(-1,1),
\end{equation}
\begin{equation}
t_j^{\mathrm{gau}}=\mu_x+\Phi^{-1}(p_j),
\end{equation}
\begin{equation}
t_j^{\mathrm{lap}}=
\mu_x+
\begin{cases}
b\log(2p_j), & p_j<0.5,\\
-b\log(2(1-p_j)), & p_j\ge 0.5,
\end{cases}
\qquad b=1/\sqrt{2},
\end{equation}
\begin{equation}
t_j^{\mathrm{log}}=\mu_x+s\log\frac{p_j}{1-p_j},
\qquad s=\sqrt{3}/\pi.
\end{equation}
Thus, linear spacing uses a bounded constant-gap template, Gaussian uses standard-normal quantiles, and Laplace/logistic use heavier-tailed quantile families under the same nominal scale convention.

\paragraph{What properties are needed.}
The diagnosis in Section~3.2 suggests that the reference geometry should first satisfy several basic desiderata:
\begin{itemize}
\item \emph{anti-collapse}: it must assign a nonzero cost to degenerate or nearly constant groups;
\item \emph{symmetry}: it should not a priori favor the top or bottom tail of a prompt group;
\item \emph{smooth rank progression}: adjacent quantiles should vary gradually rather than producing artificial hard edges;
\item \emph{fixed-scale control}: the target spread should be controlled without introducing additional shape hyperparameters.
\end{itemize}

\paragraph{Proposition 10: Gaussian induces an intermediate quantile-gap profile.}
For the purpose of comparing rank-gap profiles, we view each discrete template as the sampling of a continuous centered quantile function $Q_{\psi}(p)$ on $p\in(0,1)$, and define its local gap profile by
\begin{equation}
g_{\psi}(p)=\frac{d}{dp}Q_{\psi}(p),\qquad p\in(0,1).
\end{equation}
Under the natural continuous interpolation $Q_{\mathrm{lin}}(p)=2p-1$ for the linear template used in GRGC,
\begin{equation}
g_{\mathrm{lin}}(p)=2,
\end{equation}
\begin{equation}
g_{\mathrm{gau}}(p)=\frac{1}{\phi(\Phi^{-1}(p))},
\end{equation}
\begin{equation}
g_{\mathrm{lap}}(p)=
\begin{cases}
\dfrac{b}{p}, & p<0.5,\\
\dfrac{b}{1-p}, & p>0.5,
\end{cases}
\qquad b=1/\sqrt{2},
\end{equation}
\begin{equation}
g_{\mathrm{log}}(p)=\frac{s}{p(1-p)},
\qquad s=\sqrt{3}/\pi.
\end{equation}
Hence linear spacing is constant across all ranks, while Laplace and logistic allocate increasingly large spacing to extreme ranks at rate $\Theta((1-p)^{-1})$ as $p\to 1$ (and symmetrically as $p\to 0$). By contrast, the Gaussian gap profile also increases toward the tails but more moderately, with
\begin{equation}
g_{\mathrm{gau}}(p)=\Theta\!\left(\frac{1}{(1-p)\sqrt{\log(1/(1-p))}}\right)
\qquad \text{as } p\to 1.
\end{equation}
Thus Gaussian induces a nonconstant and smooth rank progression, but is less tail-aggressive than heavier-tailed alternatives.

\paragraph{Proof sketch.}
The formulas follow by differentiating the corresponding centered quantile functions. The asymptotic statement for the Gaussian case uses the standard tail expansion $1-p\asymp \phi(z)/z$ for $z=\Phi^{-1}(p)\to\infty$, which gives $\phi(\Phi^{-1}(p))\asymp (1-p)\sqrt{\log(1/(1-p))}$ up to constants. Substituting this into $g_{\mathrm{gau}}(p)=1/\phi(\Phi^{-1}(p))$ yields the stated tail order.

\paragraph{Interpretation.}
Proposition 10 makes precise why Gaussian is a better conditioning prior than either linear spacing or heavier-tailed alternatives. Linear targets are too rigid: they enforce the same gap everywhere and cannot distinguish central ambiguity from extreme ranks. Laplace and logistic are too tail-aggressive: they allocate an increasingly large share of the total separation budget to the extremes. Gaussian lies in the middle. It preserves nontrivial rank variation and smooth interior-to-tail progression without over-concentrating separation in the tails. This is exactly the geometry needed for conditioning rather than aggressive sharpening.

\paragraph{Why Gaussian fits these desiderata well.}
The Gaussian quantile template satisfies these basic requirements, but these requirements alone do not make it unique. Logistic and Laplace quantiles are also symmetric, non-degenerate, closed-form, and single-scale. The deciding factor is the rank-gap profile formalized in Proposition~10. Linear spacing is too flat because it assigns the same gap to every rank pair; Laplace and logistic are too tail-aggressive because their quantile gaps grow at rate $\Theta((1-p)^{-1})$ near the extremes. Gaussian quantiles change smoothly with rank while using a milder tail expansion, so they provide a balanced conditioning geometry: central candidates remain moderately separated, tails receive additional but not excessive spacing, and the template keeps a controlled scale.

\paragraph{Why simpler or sharper alternatives are less attractive.}
A uniform-quantile target would also prevent collapse, but it imposes constant spacing across the entire group. This creates a geometry with no distinction between central ambiguity and extreme ranks. In contrast, our diagnosis suggests that many groups contain several near-tied middle candidates together with a few more separated extremes; a Gaussian template accommodates this more naturally. Heavy-tailed priors such as Laplace or logistic are also possible and satisfy many of the same basic desiderata, but they allocate relatively more of the separation budget to extreme quantiles. Since our critic-side objective is meant to improve conditioning rather than aggressively enlarge extremes, the Gaussian prior is a better default because its tail progression sits between the overly rigid linear template and the heavier-tailed alternatives.

\paragraph{Why this is not generic moment regularization.}
The critic-side OT term constrains the \emph{ordered geometry} of each prompt group, not just a few aggregate statistics. Simpler penalties such as variance regularization, moment matching, or entropy-style smoothing could encourage a broader spread on average, but they would still leave many rank configurations admissible. In contrast, the OT loss ties each ordered score $r_{(j)}$ to an ordered reference quantile $t_j$, so it simultaneously constrains collapse, tail imbalance, and the progression of gaps across ranks. This is the key distinction from generic reward normalization or scalar variance control: CGC regularizes the \emph{full one-dimensional prompt-group shape} that will later be consumed by grouped optimization.

\paragraph{Summary of the Gaussian choice.}
The Gaussian target is therefore not used as a distributional assumption on semantic rewards. It is used as a group-centered ordered template whose gap profile sits between constant linear spacing and heavy-tailed alternatives. The important distinction is not that Gaussian is the only closed-form symmetric single-scale option, but that it provides the most suitable conditioning profile among the tested options: non-degenerate spread, symmetric treatment of ranks, smooth gap progression, and moderate rather than excessive tail expansion.

\subsection{Why \texttt{group\_power} Among Alternative Reward Transforms}
Our implementation contains several reward-transform alternatives, including global transforms (\texttt{linear}, \texttt{linear\_fixed}, \texttt{beta}, \texttt{beta\_logit}, \texttt{temperature}), group-aware standardization-only transforms (\texttt{group\_zscore}), stronger nonlinear transforms (\texttt{group\_sinh}), sparse winner-focused transforms (\texttt{top1\_margin\_boost}), and rank-replacement transforms (\texttt{rank\_gaussian}). The comparison is organized around the structural requirements of pre-GRPO advantage construction: the transform should be prompt-group aware, preserve critic ordering, retain magnitude information, and provide controlled gain to informative margins.

\paragraph{Concrete forms of the compared transforms.}
For clarity, the alternatives in Table~\ref{tab:ablation_PGM} correspond to the following reward-shaping rules. The proposed transform is
\begin{equation}
z_i=\frac{r_i-\mu_x}{\sigma_x+\varepsilon},
\qquad
\hat r_i^{\mathrm{power}}=\mathrm{sign}(z_i)\,|z_i|^\gamma.
\end{equation}
The \emph{Top-1} baseline corresponds to the code-level transform \texttt{top1\_margin\_boost}: if $i^\star=\arg\max_i r_i$ and $\Delta_x=r_{(N)}-r_{(N-1)}$ for increasingly sorted rewards, then only the winner is modified,
\begin{equation}
\hat r_i^{\mathrm{top1}}=
\begin{cases}
r_i+\lambda_{\mathrm{top1}}\Delta_x, & i=i^\star,\\
r_i, & \text{otherwise}.
\end{cases}
\end{equation}
The \emph{Rank} baseline corresponds to the code-level transform \texttt{rank\_gaussian}: responses are sorted within each prompt group and then reassigned Gaussian quantiles according to rank,
\begin{equation}
\hat r_{(j)}^{\mathrm{rank}}=\tau\,\Phi^{-1}(p_j),\qquad j=1,\dots,N,
\end{equation}
where $\tau$ is a scale constant. Finally, \emph{CDF} denotes a quantile-remapping baseline. Let $\hat F_x$ be the empirical CDF of the standardized scores within prompt group $x$, and let $F^{-1}$ be a fixed target inverse CDF, such as the inverse Gaussian CDF. The transform is
\begin{equation}
\hat r_i^{\mathrm{cdf}} = F^{-1}\!\left(\hat F_x(z_i)\right).
\end{equation}
This baseline preserves ordering and maps candidates to a smooth target quantile scale, but it relies mainly on empirical CDF position and therefore weakens the critic's original magnitude margins.

\paragraph{Global transforms are not prompt-group aware.}
The failures we target are groupwise: low dispersion, weak top-group margins, and unstable within-group scale. Global transforms apply the same scalar map to all samples regardless of prompt-group membership. They can rescale the critic output globally, but they do not directly repair a nearly tied group or strengthen margins relative to that group's own scale. As a result, they are mismatched to the grouped nature of the optimization problem.

\paragraph{Standardization alone is not enough.}
\texttt{group\_zscore} removes within-group shift and scale, but it does not further separate ambiguous candidates. After critic-side OT calibration, this can still leave grouped margins too weak for efficient policy updates. In other words, \texttt{group\_zscore} improves comparability but not decisiveness.

\paragraph{More aggressive transforms distort the signal more strongly.}
\texttt{group\_sinh} is group-aware and monotone, but its tails grow exponentially, which makes it more sensitive to moderately large spurious standardized scores. \texttt{top1\_margin\_boost} is highly local: it only changes the group winner and tends to create a winner-take-all geometry rather than improving the whole reward group. CDF remapping and \texttt{rank\_gaussian} both replace much of the original magnitude structure with quantile position; this preserves rank, but it weakens the critic's own margin information.

\paragraph{Why \texttt{group\_power} is the right compromise.}
The transform
\begin{equation}
\hat r_i=\mathrm{sign}(z_i)|z_i|^\gamma
\end{equation}
has four properties that are jointly important in our setting:
\begin{itemize}
\item it is \emph{group-aware}, because it operates on per-group standardized scores;
\item it is \emph{order-preserving}, because it is strictly monotone in $z_i$;
\item it is \emph{margin-amplifying}, because deviations farther from the group center receive larger polynomial local gain;
\item it is \emph{less destructive} than rank replacement and typically less brittle than exponential tails.
\end{itemize}
Thus, among the available alternatives, \texttt{group\_power} provides the most suitable compromise between preserving critic-implied ordering, sharpening already informative grouped margins, and avoiding the magnitude loss induced by quantile remapping or the overly sharp tails induced by stronger nonlinear transforms.

\paragraph{Proposition 11: signed power is the unique normalized homogeneous transform.}
The structural requirements above can be made precise. A pre-GRPO transform on standardized grouped scores should satisfy four conditions. First, it should be odd, $f(-z)=-f(z)$, because $z=0$ is the group mean after standardization and positive and negative deviations should be treated symmetrically. Second, it should be order-preserving, because the critic-implied ranking should not be reversed during advantage construction. Third, it should be positively homogeneous of degree $\gamma>1$,
\begin{equation}
f(cz)=c^\gamma f(z),\qquad c>0,
\end{equation}
so the transform applies one consistent magnitude-dependent gain rule without introducing additional shape parameters. Fourth, it should be normalized by $f(1)=1$, which fixes the arbitrary output scale and leaves $\gamma$ as the only shaping parameter.

Under these requirements, the transform is uniquely determined:
\begin{equation}
f(z)=\mathrm{sign}(z)|z|^\gamma.
\end{equation}
Moreover, for $z\ne 0$,
\begin{equation}
f'(z)=\gamma |z|^{\gamma-1},
\end{equation}
so the local gain is magnitude-dependent and grows polynomially rather than exponentially.

\paragraph{Proof sketch.}
For $z>0$, positive homogeneity and $f(1)=1$ give
\begin{equation}
f(z)=f(z\cdot 1)=z^\gamma f(1)=z^\gamma.
\end{equation}
For $z<0$, odd symmetry gives
\begin{equation}
f(z)=-f(-z)=-(-z)^\gamma=-|z|^\gamma.
\end{equation}
Thus $f(z)=\mathrm{sign}(z)|z|^\gamma$. Differentiating on $z\ne 0$ gives $f'(z)=\gamma |z|^{\gamma-1}$.

\paragraph{Interpretation.}
Proposition 11 explains why group power is not an arbitrary monotone map. If we require the transform to preserve critic ordering, treat positive and negative standardized deviations symmetrically, avoid extra shape parameters, and apply a scale-consistent power-law gain to informative margins, then the signed power form is forced. The compared alternatives violate at least one of these requirements: rank-Gaussian and CDF remapping replace magnitude information with quantile position, top-1 boosting is not a pointwise transform on the whole group, and $\sinh$ is not homogeneous and has exponentially growing local gain. This is why group power is the appropriate PGM transform for advantage construction.

\paragraph{How the two components interact.}
The two modules act on different failure modes and therefore should not be viewed as interchangeable reward shaping steps. CGC first reduces under-dispersion and scale inconsistency in the raw critic rewards $\{r_i\}$, which makes grouped normalization better conditioned but does not by itself impose a magnitude-dependent gain on already informative standardized gaps. PGM then acts on the standardized grouped structure that remains: it preserves ordering while enlarging non-near-zero gaps in the signal used to construct final advantages. Put differently, CGC improves the \emph{input conditioning} of grouped optimization, whereas PGM improves the \emph{separability} of the signal consumed after that conditioning step. This division of labor is why the full method is stronger than either component alone in Table~\ref{tab:epoch_auto_eval}.

\section{Symbol Description}
\label{a3}
To enhance clarity, a detailed description of mathematical symbols used in the present study is provided in Table~\ref{tab:symbols_description}.

\begin{table}[tbp]
    \centering
    \caption{Descriptions of symbols used in the main text and appendix.}
    \label{tab:symbols_description}
    \begin{tabular}{c|l}
    \toprule
    Symbol & Definition \\
    \midrule
    $x$ & Input prompt. \\
    $y$ & Teacher response. \\
    $\pi_T$ & Teacher policy. \\
    $\pi_\theta$ & Student policy. \\
    $\theta$ & Student-policy parameters. \\
    $G(x)_i$  & $i$-th student response. \\
    $N$ & Prompt-group size. \\
    $D(\cdot)$ & Critic / discriminator score. \\
    $r_i$ & Raw student sequence reward. \\
    $\hat r_i$ & Transformed reward. \\
    $A_i$ & GRPO advantage. \\
    $\mu_x$ & Group mean of rewards. \\
    $\sigma_x$ & Group standard deviation of rewards. \\
    $r_{(j)}$ & $j$-th ordered reward. \\
    $\hat{\mu}_x$ & Empirical reward distribution. \\
    $\nu_x$ & Continuous Gaussian reference distribution. \\
    $\nu_x^{(N)}$ & Discrete Gaussian-quantile reference distribution. \\
    $W_2^2(\hat{\mu}_x,\nu_x^{(N)})$ & Squared 2-Wasserstein distance to the discrete reference. \\
    $t_j$ & $j$-th Gaussian target quantile. \\
    $p_j$ & Rank quantile level $(j-0.5)/N$. \\
    $\Phi^{-1}(\cdot)$ & Inverse standard normal CDF. \\
    $\mathcal{L}_{\mathrm{BT}}(x)$ & Groupwise BT loss. \\
    $\mathcal{L}_{\mathrm{OT}}(x)$ & Groupwise OT loss. \\
    $\mathcal{L}_{\mathrm{critic}}(x)$ & Groupwise critic loss. \\
    $\lambda_{\mathrm{OT}}$ & OT loss weight. \\
    $z_i$ & Standardized group score. \\
    $\varepsilon$ & Numerical stability constant. \\
    $\gamma$ & Power exponent. \\
    $f_\gamma(z)$ & Signed power map. \\
    $u_i$ & Centered reward. \\
    $u_{(j)}$ & $j$-th ordered centered reward. \\
    $q_j$ & Centered Gaussian quantile. \\
    $\sigma_q$ & Target quantile standard deviation. \\
    $\rho$ & Minimum standardized magnitude in the PGM margin bound. \\
    $\rho_x$ & Empirical correlation between centered rewards and target quantiles. \\
    $z_i^\star$ & Latent standardized score. \\
    $\zeta$ & Additive reward perturbation vector. \\
    $\xi_i$ & Additive score perturbation. \\
    $\eta$ & Perturbation bound. \\
    $\eta_{\mathrm{pg}}$ & Step size in the grouped softmax surrogate update. \\
    $\Delta_{ij}^\star$ & Latent margin $z_i^\star-z_j^\star$. \\
    $\psi_j$ & Generic target-quantile template. \\
    $\tilde t_j$ & Generic target quantile. \\
    $s$ & Generic target scale. \\
    $F$ & Smooth CDF used by the CDF transform. \\
    $\Delta_x$ & Top-1 reward gap used by the Top-1 transform. \\
    $\lambda_{\mathrm{top1}}$ & Top-1 margin-boost coefficient. \\
    $\tau$ & Scale of the rank-Gaussian transform. \\
    $\langle u,q\rangle_N$ & Empirical inner product. \\
    $B$ & Responses per minibatch. \\
    $G$ & Number of prompt groups in a minibatch. \\
    $L$ & Average response length. \\
    $\mathcal{F}_{\pi}^{\mathrm{roll}}(L)$ & Actor rollout cost. \\
    $\mathcal{F}_{\pi}^{\mathrm{upd}}(L)$ & Actor update cost. \\
    $\mathcal{F}_{D}(L)$ & Critic pass cost. \\
    $\mathcal{M}_{\pi}$ & Actor memory cost. \\
    $\mathcal{M}_{D}$ & Critic memory cost. \\
    \bottomrule
    \end{tabular}
\end{table}

\section{Training Pseudocode}
\label{app:pseudocode}
Algorithm~\ref{alg:grgc} summarizes the full GRGC training pipeline in the same spirit as the GAD pseudocode. We separate the warmup phase from the on-policy GRGC phase, and explicitly show where critic-side OT calibration and policy-side group power modulation are applied.

\begin{algorithm}[tb]
\small
\caption{\ours{}: Groupwise Reward Geometry Conditioning}
\label{alg:grgc}
\begin{algorithmic}
\Require Distillation data $\mathcal{T}=\{(x,y)\}$; student policy $\pi_\theta$; critic $D$; OT weight $\lambda_{\mathrm{OT}}$; power exponent $\gamma$; group size $N$
\Ensure Trained student policy $\pi_\theta$
\Statex
\State \textbf{\textit{Warmup Stage}}
\For{each batch $(x,y)\sim\mathcal{T}$}
    \State Update student $\pi_\theta$ on teacher responses $y$ with cross-entropy loss
    \State Sample a prompt-wise student group $\{G(x)_i\}_{i=1}^N$ from $\pi_\theta$
    \State Compute critic scores $D(y)$ and $\{r_i=D(G(x)_i)\}_{i=1}^N$
    \State Compute teacher-student BT loss $\mathcal{L}_{\mathrm{BT}}(x)$
    \State Construct unit-scale Gaussian target quantiles $\{t_j\}_{j=1}^N$ with group mean $\mu_x$
    \State Compute OT loss $\mathcal{L}_{\mathrm{OT}}(x)=\frac{1}{N}\sum_{j=1}^{N}(r_{(j)}-t_j)^2$
    \State Update critic $D$ with $\mathcal{L}_{\mathrm{critic}}(x)=\mathcal{L}_{\mathrm{BT}}(x)+\lambda_{\mathrm{OT}}\mathcal{L}_{\mathrm{OT}}(x)$
\EndFor
\Statex
\State \textbf{\textit{\ours{} Training Stage}}
\Repeat
    \For{each batch $(x,y)\sim\mathcal{T}$}
        \State Sample a prompt-wise student group $\{G(x)_i\}_{i=1}^N$ from $\pi_\theta$
        \State Compute critic scores $D(y)$ and $\{r_i=D(G(x)_i)\}_{i=1}^N$
        \State Compute teacher-student BT loss $\mathcal{L}_{\mathrm{BT}}(x)$
        \State Construct unit-scale Gaussian target quantiles $\{t_j\}_{j=1}^N$ with group mean $\mu_x$
        \State Compute OT loss $\mathcal{L}_{\mathrm{OT}}(x)=\frac{1}{N}\sum_{j=1}^{N}(r_{(j)}-t_j)^2$
        \State Update critic $D$ with $\mathcal{L}_{\mathrm{critic}}(x)=\mathcal{L}_{\mathrm{BT}}(x)+\lambda_{\mathrm{OT}}\mathcal{L}_{\mathrm{OT}}(x)$
        \State Standardize raw rewards within each prompt group to obtain $\{z_i\}_{i=1}^N$
        \State Apply group power shaping $\hat r_i=\mathrm{sign}(z_i)|z_i|^\gamma$
        \State Construct GRPO advantages $\{A_i\}_{i=1}^N$ from transformed rewards $\{\hat r_i\}_{i=1}^N$
        \State Update student $\pi_\theta$ with the GRPO objective using $\{A_i\}_{i=1}^N$
    \EndFor
\Until{convergence}
\State \Return $\pi_\theta$
\end{algorithmic}
\end{algorithm}

\section{Automatic Evaluation Details}
\label{app:eval_detail}
We use greedy decoding and set the maximum response length to 1536 tokens, except for the Doubao-Seed-2.0 LMSYS setting where responses are capped at 2048 tokens as described in the implementation details. Prompts are constructed using the same prompt wrapper as MiniLLM~\cite{gu2024minillm} and GAD~\cite{gad}, which is illustrated in Figure~\ref{fig:prompt_wrapper} and Figure~\ref{fig:prompt_gpt4}. For evaluation with Qwen2.5-72B feedback, we first use Qwen2.5-72B to generate a reference answer for each instruction using the prompt in Figure~\ref{fig:prompt_wrapper}. We then present the student response and the corresponding Qwen2.5-72B reference answer to the judge model, which evaluates both responses using the prompt shown in Figure~\ref{fig:prompt_gpt4}.

The judge assigns one scalar score to the student response and one scalar score to the reference response. The reported automatic evaluation score is computed as
\begin{equation}
\mathrm{Score}
=
\frac{\mathrm{StudentScore}}
{\mathrm{StudentScore} + \mathrm{RefScore}},
\end{equation}
where $\mathrm{StudentScore}$ denotes the judge score assigned to the student output and $\mathrm{RefScore}$ denotes the judge score assigned to the paired reference answer. This normalized score lies in $(0,1)$ and measures the quality of the student response relative to the reference answer. In the reported experimental metrics, we report $100 \times \mathrm{Score}$, so the automatic evaluation results are presented on a $0$ to $100$ scale.

In addition to the averaged score, we also report \emph{win rate} when appropriate. For a given evaluation set, a sample is counted as a \emph{win} if the judge assigns a strictly higher scalar score to the student response than to the reference response, a \emph{loss} if the student score is strictly lower, and a \emph{tie} otherwise. The win rate is then defined as
\begin{equation}
\mathrm{WinRate}
=
\frac{\#\{\mathrm{StudentScore} > \mathrm{RefScore}\}}
\#\{\text{all evaluated samples}\}.
\end{equation}
Thus, win rate measures the fraction of evaluation prompts on which the student is preferred to the reference answer under the same judge, while the averaged score reflects the relative quality margin after normalization.

\begin{figure}[h]
    \begin{tcolorbox}
    Below is an instruction that describes a task. \\
    Write a response that appropriately completes the request. \\ \\
    \#\#\# Instruction: \\
    \{instruction\} \\ \\
    \#\#\# Response:
    \end{tcolorbox}
    \caption{The prompt wrapper for training and evaluation.}
    \label{fig:prompt_wrapper}
\end{figure}

\begin{figure}[h]
    \begin{tcolorbox}
        We would like to request your feedback on the performance of two AI assistants in response to the user instruction and input displayed above. \\
        Please rate the helpfulness, relevance, accuracy, and level of detail of their responses. Each assistant receives an overall score on a scale of 1 to 10, where a higher score indicates better overall performance. \\
        Please first output a single line containing only two values indicating the scores for Assistant 1 and 2, respectively. The two scores are separated by a space. \\
        In the subsequent line, please provide a comprehensive explanation of your evaluation, avoiding any potential bias and ensuring that the order in which the responses were presented does not affect your judgment.
    \end{tcolorbox}
    \caption{Automatic evaluation prompt.}
    \label{fig:prompt_gpt4}
\end{figure}

\section{Further Experimental Analyses}
\label{a4}

\subsection{Sensitivity Analysis}
GRGC requires tuning only two method-specific hyperparameters beyond the underlying GAD training recipe: the critic-side OT weight $\lambda_{\mathrm{OT}}$ and the policy-side power exponent $\gamma$. The Gaussian reference uses unit-scale quantiles throughout our experiments. We therefore focus the sensitivity study on $\lambda_{\mathrm{OT}}$ and $\gamma$. As shown in Table~\ref{tab:sens_lambda_ot} and Table~\ref{tab:sens_gamma}, the method is stable across a reasonably broad range, which is important in practice because it means GRGC does not require delicate tuning to outperform baseline GAD.

Table~\ref{tab:sens_lambda_ot} studies the OT weight $\lambda_{\mathrm{OT}}$. When $\lambda_{\mathrm{OT}}$ is too small, the critic-side geometry constraint becomes weak and the method moves toward a lightly regularized GAD regime, so under-dispersion and unstable reward scale are only partially corrected. When $\lambda_{\mathrm{OT}}$ is too large, the critic becomes overly conservative and policy improvement slows down because the reward geometry is over-regularized. Between these two extremes, however, performance is very stable: values from roughly $0.005$ to $0.05$ give nearly identical results, and we therefore adopt $\lambda_{\mathrm{OT}}=0.01$ as a simple default.

Table~\ref{tab:sens_gamma} studies the power exponent $\gamma$. When $\gamma$ is too close to $1$, the transform degenerates toward a near-identity mapping, so the GRPO-side shaping effect becomes weak and only part of the grouped-margin problem is recovered. When $\gamma$ is too large, the transformed rewards become overly sharp and the grouped optimization signal becomes less robust. Again, a broad intermediate region remains stable, with values from about $1.2$ to $1.6$ producing very similar results. For convenience, we use $\gamma=1.5$ throughout the paper.

Overall, these results show that GRGC is easy to tune in practice: only two intuitive parameters are tuned, both admit stable operating ranges rather than narrow optima, and the default choices $(\lambda_{\mathrm{OT}}=0.01,\gamma=1.5)$ work well across model scales. This makes the method convenient to deploy in practical black-box distillation settings.

\begin{table}[]
    \centering
        \caption{Sensitivity to the critic-side OT weight $\lambda_{\mathrm{OT}}$. Results are averaged Qwen2.5-72B evaluation scores. Very small $\lambda_{\mathrm{OT}}$ weakens geometry conditioning, while overly large $\lambda_{\mathrm{OT}}$ makes the critic too conservative. A broad middle range remains stable, and we use $\lambda_{\mathrm{OT}}=0.01$ in the main experiments.}
    \label{tab:sens_lambda_ot}
    \begin{tabular}{c|cccccc}
     \toprule
      $\lambda_{\text{OT}}$ & 0.001 & 0.005 & 0.01 & 0.02 & 0.05 & 0.1\\
      \midrule
      Qwen2.5-3B-Instruct &48.92 &50.10&50.19&50.22&50.16&49.94 \\
      Qwen2.5-1.5B-Instruct &46.79 &47.27&47.36&47.31&47.32&47.17 \\
          \bottomrule
    \end{tabular}

\end{table}
    \begin{table}[]
    \centering
        \caption{Sensitivity to the power exponent $\gamma$. Results are averaged Qwen2.5-72B evaluation scores. Values too close to $1$ reduce the shaping effect, whereas overly large values lead to overly sharp grouped rewards. A broad intermediate region is stable, and we use $\gamma=1.5$ in the main experiments.}
    \label{tab:sens_gamma}
    \begin{tabular}{c|cccccc}
     \toprule
    $\gamma$ & 1.0 & 1.2 & 1.4 & 1.5 & 1.6 & 2.0 \\ \midrule
      Qwen2.5-3B-Instruct &47.52 &50.20&50.15&50.19&50.14&49.72 \\
      Qwen2.5-1.5B-Instruct &46.40 &47.29&47.34&47.36&47.30&47.02 \\
    \bottomrule
    \end{tabular}

\end{table}

\subsection{Distillation with an Explicit Step-by-Step Teacher Instruction}
We additionally evaluate our framework in a more challenging Doubao-Seed-2.0 teacher setting by adding an explicit ``think step by step'' instruction to the input prompt. Doubao-Seed-2.0 uses thinking mode in both this setting and the standard setting; the difference here is the additional prompt instruction. This modification encourages the teacher to produce longer and more structured responses, making the resulting supervision more demanding than standard concise instruction-following outputs. Table~\ref{tab:autoapp} summarizes the results. Compared with the standard Doubao-Seed-2.0 teacher responses in Table~\ref{tab:auto_dolly}, the step-by-step teacher achieves higher win rates against the reference answers on LMSYS, Dolly, and SelfInst, increasing from $82.5\%/75.8\%/74.0\%$ to $83.1\%/82.4\%/86.4\%$, while remaining equally strong on Vicuna at $93.8\%$.

The student results show that this stronger teacher behavior is also transferred more effectively. This indicates that when the teacher provides higher-quality and more structured answers, GRGC can better translate the richer supervision into stronger student responses by improving the grouped reward signal used for policy optimization.

Methodologically, this setting is useful because it enlarges the gap between simple sequence imitation and on-policy optimization from grouped rewards. When teacher answers become longer and structurally richer, merely copying surface form becomes less sufficient, and the quality of grouped reward signals becomes more important. The step-by-step teacher experiments therefore provide an additional stress test showing that GRGC remains effective when the teacher is encouraged to produce richer reasoning-style supervision.
\begin{table}[t]
\caption{Automatic evaluation results for models distilled from Doubao-Seed-2.0 with an explicit step-by-step instruction and trained on the Dolly training set. We report the averaged Qwen2.5-72B evaluation score on the test datasets.}
\small
\centering
\resizebox{\linewidth}{!}{
\begin{tabular}{cc|cc|cc|cc|cc}
\toprule
\multirow{2}{*}{Model} & \multirow{2}{*}{Method}
& \multicolumn{2}{c|}{LMSYS}
& \multicolumn{2}{c|}{Dolly}
& \multicolumn{2}{c|}{SelfInst}
& \multicolumn{2}{c}{Vicuna} \\
&
& Score & Win
& Score & Win
& Score & Win
& Score & Win \\
\midrule
Doubao-Seed-2.0 & Teacher
& 53.58 & 83.1\%
& 53.47 & 82.4\%
& 53.78 & 86.4\%
& 53.63 & 93.8\%  \\
\midrule

\multirow{4}{*}{Qwen2.5-3B-Instruct}
& Before Distill.
& 45.79 & 11.9\%
& 44.93 & 4.0\%
& 46.56 & 12.8\%
& 47.85 & 3.8\% \\
& SeqKD
& 46.09 & 41.8\%
& 46.03 & 48.8\%
& 46.81 & 46.3\%
& 48.39 & 66.3\% \\
& GAD
& 47.85 & 51.4\%
& 47.61 & 52.0\%
& 48.24 & 51.7\%
& 50.32 & 73.8\% \\
& \textbf{\ours}
& \bf 48.84 & \bf 55.1\%
& \bf 48.98 & \bf 61.4\% 
& \bf 49.17 & \bf 53.7\%
& \bf 51.85& \bf 82.5\% \\
\midrule

\multirow{4}{*}{Qwen2.5-1.5B-Instruct}
& Before Distill.
& 41.93 & 4.8\%
& 39.78 & 0.6\%
& 41.06 & 5.0\%
& 43.24 & 0.0\% \\
& SeqKD
& 39.91 & 19.2\%
& 41.81 & 36.0\%
& 42.07 & 29.3\%
& 45.79 & 38.8\%\\
& GAD
& 39.86 & 20.3\%
& 43.06 & 31.0\%
& 44.78 & 31.8\%
& 47.54 & 41.3\% \\
& \textbf{\ours}
& \bf 43.51 & \bf35.3\%
& \bf 44.08 & \bf 44.2\%
& \bf 45.88 & \bf39.3\%
& \bf 48.43 & \bf 53.8\% \\
\bottomrule
\end{tabular}
}
\label{tab:autoapp}
\end{table}

\subsection{Human Evaluation Results}
\begin{figure}
\centering
\includegraphics[width=\linewidth]{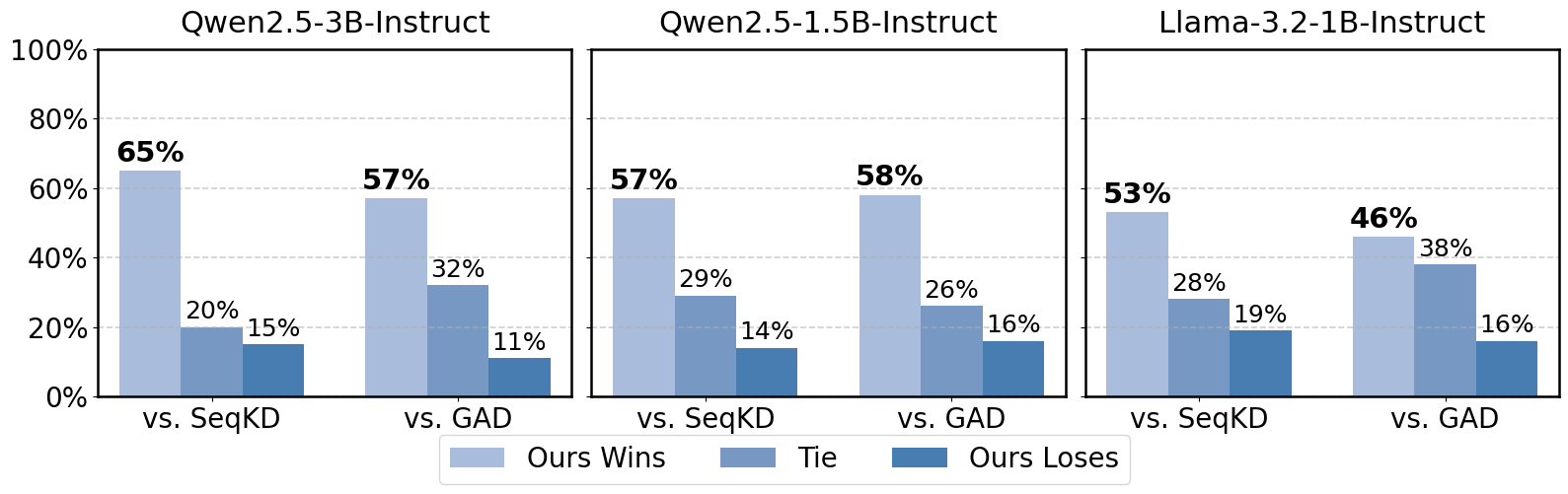}
\caption{Human evaluation results on test sets. We compare \ours{} to the models fine-tuned with SeqKD~\cite{seqkd} and GAD~\cite{gad}.}
\label{fig:user-study}
\end{figure}
Automatic metrics are useful but incomplete, so we additionally conduct a human study. We engage 10 professional volunteers for annotation. For each evaluated model, each volunteer judges 50 prompt-response groups, ensuring a balanced evaluation set that covers all four test sources used in the paper: 20 groups from the LMSYS test set, 10 from Dolly, 10 from SelfInst, and 10 from Vicuna. All annotators evaluate the same set of prompt-response groups and compare the candidate responses under the same prompt, selecting the better answer based on overall helpfulness, relevance, correctness, and completeness.

The annotators are shown the prompt together with the candidate responses and are asked to make a preference judgment using the same criteria throughout the study. The human study only involves expert evaluation of model outputs and does not collect personal or sensitive user information; no separate crowdsourcing platform is used.

Figure~\ref{fig:user-study} shows that \ours{} is preferred over both SeqKD and GAD in human comparison. This result is important because it confirms that the gains of GRGC are not limited to judge-model scores or response-format differences. Instead, the improvements induced by reward-geometry conditioning are visible to human evaluators as better overall answer quality.

\subsection{More Visualization Analysis}
\noindent
\begin{minipage}[t]{0.49\linewidth}
    \vspace{0pt}
    \centering
    \includegraphics[width=\linewidth]{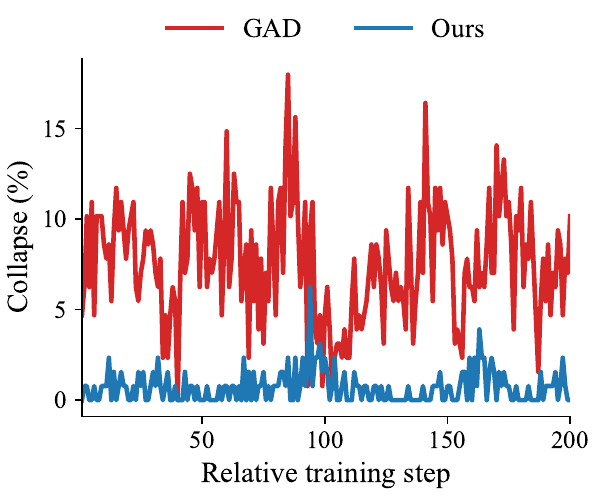}
    \captionof{figure}{Collapsed prompt groups during training. GRGC keeps the fraction with relative reward std below $0.15$ consistently lower than GAD.}
    \label{fig:ours_geometry_window}
\end{minipage}
\hfill
\begin{minipage}[t]{0.47\linewidth}
    \vspace{0pt}
    \centering
    \includegraphics[width=\linewidth]{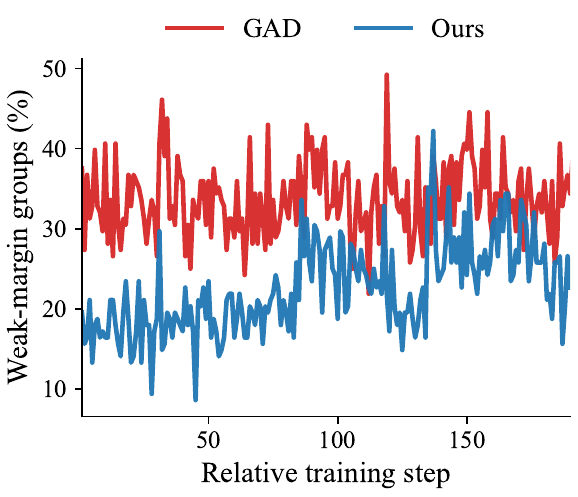}
    \captionof{figure}{Weak-margin groups during training. GRGC reduces the fraction whose normalized top-2 gap is below $0.08$.}
    \label{fig:weakmargin_baseline_vs_ours}
\end{minipage}

\begin{figure}[t]
    \centering
    \includegraphics[width=0.5\linewidth]{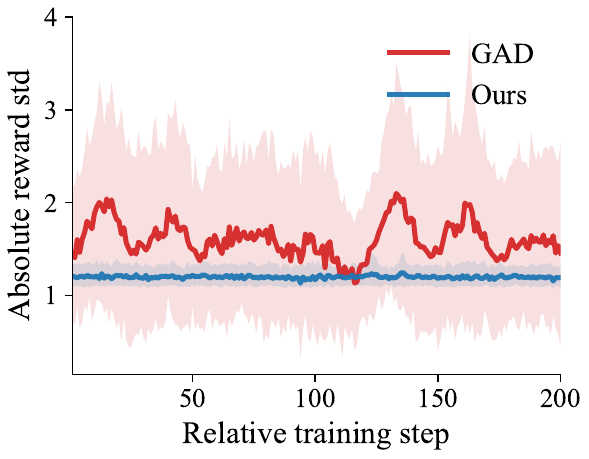}
    \caption{Unnormalized prompt-group reward spread . The solid lines show the mean prompt-group reward standard deviation, and shaded regions show the 10th--90th percentile range across prompt groups at each step. GRGC keeps the absolute reward spread in a more controlled range than GAD, confirming that the relative-spread improvement is not caused only by normalization with the global reward scale.}
    \label{fig:absolute_spread_compare}
\end{figure}

Figure~\ref{fig:ours_geometry_window} and Figure~\ref{fig:weakmargin_baseline_vs_ours} provide a more direct visualization of the geometric changes induced by GRGC. Here, a \emph{collapsed prompt group} means that the reward spread within a prompt-wise group is too small, so the candidate responses become poorly separated by the critic; concretely, we mark a group as collapsed when its relative reward standard deviation satisfies $\mathrm{rel\_std}<0.15$, and the reported collapse ratio is the fraction of such groups in a training window. A \emph{weak-margin group} means that the top candidates remain too close even after grouping. We define the top-2 margin as the reward difference between the highest-scored and second-highest-scored student responses within the same prompt group, and normalize it by the full within-group reward span to obtain $\mathrm{top1\_top2\_gap\_ratio}$. Concretely, we mark a group as weak-margin when $\mathrm{top1\_top2\_gap\_ratio}<0.08$, and the reported weak-margin ratio is the fraction of prompt groups with such insufficient top-2 separation.

Under these definitions, Figure~\ref{fig:ours_geometry_window} shows that our method suppresses collapsed prompt groups throughout the training window, rather than only improving a single summary statistic at the end of training. Figure~\ref{fig:absolute_spread_compare} complements the relative-spread plot by showing that the unnormalized prompt-group reward standard deviation is also kept in a more controlled range under GRGC, so the effect is not simply produced by dividing by the global reward scale. Figure~\ref{fig:weakmargin_baseline_vs_ours} further shows that this improvement is accompanied by a lower proportion of weak-margin groups in the reward signal eventually consumed by grouped optimization. Taken together, these visualizations support the intended division of labor between the two components of GRGC: critic-side OT calibration reduces group collapse, while policy-side reward shaping makes the remaining grouped margins more usable for GRPO.

These figures also clarify why our method should not be interpreted as merely making rewards ``larger'' or ``more extreme.'' The goal is not to maximize separation indiscriminately, but to keep prompt groups neither collapsed nor dominated by uninformative near-ties. In this regime, grouped optimization can exploit reward differences more reliably and with fewer pathological updates.

\subsection{GPT-OSS Evaluation Results}
We further evaluate the LMSYS-trained models with GPT-OSS-120B as an additional automatic judge, in order to check whether the observed gains depend on the specific Qwen2.5-72B evaluation model used in the main experiments. The evaluation follows the same automatic protocol described in Appendix~\ref{app:eval_detail}: the judge assigns scalar scores to the student response and the reference response, from which we compute the normalized score and win rate; the only change is that GPT-OSS is used as the judge model. Table~\ref{tab:gptoss} shows the same trend under this alternative evaluator. For Qwen2.5-3B-Instruct, \ours{} improves the score from $56.41$ with GAD to $60.52$, and raises the win rate from $50.1\%$ to $60.8\%$. The improvement is even larger for Qwen2.5-1.5B-Instruct, where \ours{} improves the score from $55.56$ to $59.77$ and the win rate from $47.6\%$ to $59.9\%$. These results indicate that the advantage of GRGC is not tied to a single judge model: the better reward geometry learned during adversarial distillation translates into responses that are also preferred by an independent GPT-OSS evaluator.

\begin{table}[t]
    \caption{GPT-OSS-120B evaluation results on the LMSYS test set. We report the averaged GPT-OSS-120B evaluation score and win rate.}

    \centering
    \begin{tabular}{c|cc|cc}
    \toprule
    \multirow{2}{*}{Method}
    & \multicolumn{2}{c|}{Qwen2.5-3B-Instruct}
    & \multicolumn{2}{c}{Qwen2.5-1.5B-Instruct} \\
    \cmidrule(lr){2-3}
    \cmidrule(lr){4-5}
    & Score & Win Rate
    & Score & Win Rate \\
    \midrule
     GPT-5-Chat Teacher      &66.13  &76.8\%  &66.13  &76.8\%  \\
     \midrule
    Before Distill          &54.75  &49.5\%  &52.81  &43.4\%  \\
    SeqKD~\cite{seqkd}      & 55.65 & 54.7\% & 54.20 &46.8\%  \\
    GAD~\cite{gad}          & 56.41 &50.1\%  & 55.56 &47.6\%  \\
    Ours                    & \bf 60.52 & \bf 60.8\% & \bf59.77 & \bf59.9\% \\
    \bottomrule
    \end{tabular}
    \label{tab:gptoss}
\end{table}

\subsection{Analysis of Token Length}
\begin{table}[t]
\centering
\caption{Average response length (token length) of different methods and models on the LMSYS test set. The models are trained using the LMSYS-Chat training set with the GPT-5-Chat teacher.}
\begin{tabular}{lccccc}
\toprule
Model & Teacher & Before Distill. & SeqKD & GAD & Ours \\
\midrule
Qwen2.5-3B-Instruct   & 329.1 & 338.9 & 318.2 & 438.0 & 328.1 \\
Qwen2.5-1.5B-Instruct & 329.1 & 317.8 & 310.4 & 396.1 & 335.7 \\
\bottomrule
\end{tabular}
\label{tab:token_length}
\end{table}
Table~\ref{tab:token_length} reports average response length as a behavioral diagnostic rather than as a direct quality metric. A good distilled student should avoid both severe under-generation, which often indicates missing content or truncated reasoning, and unnecessary verbosity, which often reflects repetitive or weakly controlled generation. From this perspective, \ours{} is attractive because its response length remains much closer to the teacher than GAD while avoiding the strong length inflation that appears in standard adversarial distillation. SeqKD, by contrast, tends to produce shorter responses, which is consistent with its reliance on static teacher trajectories and its tendency to under-explore during on-policy rollout.

This result fits the broader interpretation of GRGC. Reward geometry conditioning does not simply encourage the student to speak more; rather, it provides a better-conditioned optimization signal so that the student can allocate response length more appropriately. In our experiments, this translates into outputs whose lengths remain closer to the teacher's response budget while still achieving higher automatic and human-evaluation performance.

\subsection{Rule-based Instruction Following on IFEval}
We evaluate instruction following on IFEval~\cite{zhou2023ifeval}, whose verifiable constraints are scored by deterministic rule-based checkers rather than an LLM judge. Table~\ref{tab:ifeval} reports prompt-level strict and loose accuracy. GRGC achieves the best result at both student scales: relative to GAD, it improves strict/loose accuracy by $1.80/2.28$ points for Qwen2.5-1.5B and $2.63/3.48$ points for Qwen2.5-3B. These results provide judge-free evidence that the gains extend to verifiable instruction compliance.

\begin{table}[H]
\centering
\small
\caption{Rule-based IFEval accuracy (\%) under strict and loose verification.}
\label{tab:ifeval}
\begin{tabular}{llcc}
\toprule
Student & Method & Strict & Loose \\
\midrule
\multirow{3}{*}{Qwen2.5-1.5B}
& SeqKD & 54.56 & 57.79 \\
& GAD   & 54.32 & 58.51 \\
& \textbf{GRGC}  & \textbf{56.12} & \textbf{60.79} \\
\midrule
\multirow{3}{*}{Qwen2.5-3B}
& SeqKD & 66.91 & 70.14 \\
& GAD   & 68.35 & 72.06 \\
& \textbf{GRGC}  & \textbf{70.98} & \textbf{75.54} \\
\bottomrule
\end{tabular}
\end{table}

\subsection{Long-form Mathematical Reasoning}
We further test GRGC outside open-ended chat by sampling 30,000 training problems from OpenR1-Math-220k~\cite{openr1math220k}, which provides DeepSeek-R1-generated reasoning trajectories~\cite{deepseekr1}. We train Qwen2.5-1.5B with a 10,000-token maximum sequence length for teacher trajectories, student rollouts, and critic scoring; many teacher and student outputs exceed 8,000 tokens. We evaluate benchmark accuracy on Minerva Math~\cite{lewkowycz2022minerva} and 340 retained Gaokao-MathQA questions from AGIEval~\cite{zhong2024agieval}, both of which have deterministic reference answers.

Table~\ref{tab:long_math} shows that GRGC achieves the highest accuracy on both benchmarks. It improves over GAD by $2.58$ points on Minerva Math and $3.53$ points on Gaokao-MathQA, while exceeding the undistilled student by $6.25$ and $5.59$ points, respectively. The gains therefore persist under domain-specific supervision, ground-truth evaluation, and long reasoning trajectories beyond the response lengths used in our chat experiments.

\begin{table}[H]
\centering
\small
\caption{Accuracy (\%) of Qwen2.5-1.5B after long-form mathematical-reasoning distillation on 30000 OpenR1-Math-220k problems.}
\label{tab:long_math}
\begin{tabular}{lcc}
\toprule
Method & Minerva Math & Gaokao-MathQA \\
\midrule
Before Distillation & 11.40 & 43.53 \\
SeqKD               & 13.97 & 43.82 \\
GAD                 & 15.07 & 45.59 \\
\textbf{GRGC}       & \textbf{17.65} & \textbf{49.12} \\
\bottomrule
\end{tabular}
\end{table}

\subsection{Evaluation on Math500}
We additionally evaluate mathematical reasoning on Math500~\cite{math500}. This setting is intentionally challenging for our training distribution: the student is distilled from GPT-5-Chat on LMSYS-Chat, where only $3.4\%$ of the training prompts are math-related. Thus, Math500 measures whether black-box distillation preserves and transfers sparse reasoning ability from a mostly general conversational dataset, rather than whether the method is specialized for math training.

We compute Math500 accuracy with a rule-based answer-matching protocol rather than an LLM judge. Each model is prompted with the standard instruction wrapper plus an explicit request to put the final answer in \verb|\boxed{}|, and responses are generated greedily. For each output, we extract the final answer by first taking the last \verb|\boxed{}| expression, then falling back to answers after \verb|####|, explicit phrases such as ``final answer is'', and finally the last mathematical expression or number. The extracted answer is compared against the dataset's \texttt{final\_answer} field using exact string match after LaTeX normalization, numerical comparison with tolerance $10^{-6}$ for numbers and fractions, and SymPy-based symbolic equivalence when available. Extraction failures are counted as incorrect, and accuracy is the percentage of correct examples among the 500 Math500 problems.

Table~\ref{tab:math500} shows that naive sequence-level imitation can hurt mathematical reasoning: SeqKD drops from the undistilled model's $46.4\%$ accuracy to $39.8\%$. GAD partially recovers this degradation, reaching $44.0\%$, but still remains below the original student. In contrast, \ours{} achieves $49.6\%$, outperforming the undistilled model and both distillation baselines. This result is consistent with our main claim: by improving grouped advantage construction from adversarial rewards, GRGC provides a better-conditioned on-policy signal that can exploit the limited math-related supervision in LMSYS-Chat without degrading the student's reasoning ability.

\begin{table}[]
    \centering
        \caption{Math500 evaluation for Qwen2.5-1.5B-Instruct distilled from GPT-5-Chat on LMSYS-Chat. Only $3.4\%$ of LMSYS-Chat training prompts are math-related, making this an out-of-domain reasoning stress test rather than a math-specialized training setting.}
    \begin{tabular}{c|cccc}
    \toprule
      Method   & Before Distill & SeqKD & GAD& Ours \\
    \midrule
      Accuracy   &46.4\% &39.8\%&44.0\%&49.6\%\\ 
    \bottomrule
    \end{tabular}
    \label{tab:math500}
\end{table}

\subsection{Multi-seed Evaluation Robustness}
The automatic evaluation results in the main paper use the default evaluation seed, \ie, Seed=0. 
To verify that the observed gains do not come from judge-sampling randomness, we additionally repeat the Qwen2.5-72B automatic evaluation with five evaluation seeds while keeping the model generations fixed. 
Tables~\ref{seedscore} and~\ref{seedwinrate} report both the individual seed results and the mean/std across seeds. 
The results are robust across evaluation seeds: the standard deviation is only $0.03$ for our normalized score and $0.53$ for win rate, while the improvement of \ours{} over GAD is much larger than these fluctuations. 
This confirms that the performance gains of GRGC are robust to evaluation-seed variation rather than being an artifact of a particular judge sample.

\begin{table}[]
    \centering
        \caption{Multi-seed Qwen2.5-72B automatic evaluation score. Seed=0 is the default evaluation seed used in the main results. The 95\% Confidence Interval (CI) represents the range within which the true mean score is expected to lie with 95\% confidence.}
    \begin{tabular}{c|ccccccc|c}
    \toprule
      Method   & Seed=0 & Seed=1 & Seed=2 & Seed=3 & Seed=4 & Mean & Std & 95\% CI\\
    \midrule
  SeqKD & 47.06 & 47.10 & 47.05 & 47.04 & 47.07 & 47.06 & 0.02 & [47.04, 47.08] \\
GAD   & 48.44 & 48.44 & 48.43 & 48.38 & 48.43 & 48.42 & 0.03 & [48.38, 48.46] \\
Ours  & 50.19 & 50.20 & 50.15 & 50.22 & 50.18 & 50.19 & 0.03 & [50.15, 50.23] \\
    \bottomrule
    \end{tabular}
    \label{seedscore}
\end{table}

\begin{table}[]
    \centering
    \caption{Multi-seed Qwen2.5-72B automatic evaluation win rate. Seed=0 is the default evaluation seed used in the main results. The 95\% Confidence Interval (CI) represents the range within which the true mean score is expected to lie with 95\% confidence.}
    \begin{tabular}{c|ccccccc|c}
    \toprule
      Method   & Seed=0 & Seed=1 & Seed=2 & Seed=3 & Seed=4 & Mean & Std & 95\% CI\\
    \midrule
    SeqKD & 18.4 & 18.6 & 18.4 & 18.8 & 18.2 & 18.46 & 0.24 & [18.16, 18.75] \\
    GAD   & 25.1 & 25.5 & 24.6 & 24.8 & 24.8 & 24.97 & 0.32 & [24.58, 25.36] \\
    Ours  & 45.3 & 45.3 & 44.9 & 45.3 & 46.3 & 45.43 & 0.54 & [44.75, 46.10] \\
    \bottomrule
    \end{tabular}
    \label{seedwinrate}
\end{table}

\section{Complexity Analysis}
\label{a5}
Let $B$ be the total number of student responses in a minibatch, $N$ the prompt-group size, and $G=B/N$ the number of prompt groups. Let $L$ denote the average generated response length, and let $\mathcal{F}_{\pi}^{\mathrm{roll}}(L)$, $\mathcal{F}_{\pi}^{\mathrm{upd}}(L)$, and $\mathcal{F}_{D}(L)$ denote the model-scale costs of actor rollout, actor update, and critic forward-backward on a length-$L$ response. This notation separates expensive neural-network computation from cheap grouped scalar operations such as BT loss, GRPO normalization, OT calibration, and group power shaping.

\paragraph{Baseline GAD.}
For one minibatch, baseline GAD contains four conceptually distinct pieces:
\begin{enumerate}
\item \textbf{Actor rollout.} Generating $B$ student responses of average length $L$ costs
\begin{equation}
\mathcal{O}\!\big(B\,\mathcal{F}_{\pi}^{\mathrm{roll}}(L)\big).
\end{equation}
\item \textbf{Critic scoring and BT training.} The critic evaluates the $G$ teacher responses and the $B$ student responses, then forms Bradley--Terry losses on the resulting sequence scores. Thus the exact critic-side cost is
\begin{equation}
\mathcal{O}\!\big((B+G)\,\mathcal{F}_{D}(L)\big) + \mathcal{O}(B),
\end{equation}
where the extra $\mathcal{O}(B)$ term is the scalar BT loss itself after sequence scores are formed. Since $G=B/N$ and $N$ is fixed in our setting, this simplifies to
\begin{equation}
\mathcal{O}\!\big(B\,\mathcal{F}_{D}(L)\big) + \mathcal{O}(B).
\end{equation}
\item \textbf{GRPO normalization.} Groupwise mean/std computation and advantage construction operate only on $B$ scalar sequence rewards:
\begin{equation}
\mathcal{O}(B).
\end{equation}
\item \textbf{Actor update.} PPO/GRPO updates over the generated trajectories cost
\begin{equation}
\mathcal{O}\!\big(B\,\mathcal{F}_{\pi}^{\mathrm{upd}}(L)\big).
\end{equation}
\end{enumerate}
Hence the total time complexity of GAD can be written as
\begin{equation}
\mathcal{C}_{\mathrm{GAD}}
=
\mathcal{O}\!\big(B\,\mathcal{F}_{\pi}^{\mathrm{roll}}(L)\big)
+
\mathcal{O}\!\big(B\,\mathcal{F}_{D}(L)\big)
+
\mathcal{O}\!\big(B\,\mathcal{F}_{\pi}^{\mathrm{upd}}(L)\big)
+
\mathcal{O}(B),
\end{equation}
where the final $\mathcal{O}(B)$ term absorbs BT loss formation and GRPO normalization. In other words, the dominant cost of GAD comes from rollout, critic forward-backward, and actor update; BT loss and GRPO normalization themselves are low-order grouped scalar operations.

\paragraph{Critic-side OT calibration.}
Our OT term is added after the critic has already produced one scalar sequence reward for each student response. For each prompt group, OT sorts the $N$ scalar rewards and matches them to $N$ Gaussian target quantiles. Sorting dominates this step, yielding
\begin{equation}
\mathcal{O}(N \log N)
\end{equation}
time per prompt group and therefore
\begin{equation}
\mathcal{O}(B \log N)
\end{equation}
time per minibatch. The additional memory cost is linear in the number of grouped scalar rewards:
\begin{equation}
\mathcal{O}(B).
\end{equation}
Because $N$ is small and fixed in our experiments (e.g., $N=8$), this overhead is low-order relative to the critic forward-backward pass itself.

\paragraph{Policy-side group power modulation.}
The group power transform computes a per-group mean and standard deviation, standardizes the grouped rewards, and applies an elementwise signed power map. These are all scalar operations on $B$ rewards:
\begin{equation}
\mathcal{O}(N)
\end{equation}
per prompt group and
\begin{equation}
\mathcal{O}(B)
\end{equation}
per minibatch, with
\begin{equation}
\mathcal{O}(B)
\end{equation}
additional memory. Thus group power is asymptotically cheaper than OT and negligible relative to rollout or model updates.

\paragraph{Overall comparison with GAD.}
Combining the two additions gives
\begin{equation}
\mathcal{C}_{\mathrm{GRGC}}
=
\mathcal{C}_{\mathrm{GAD}} + \mathcal{O}(B \log N) + \mathcal{O}(B)
=
\mathcal{C}_{\mathrm{GAD}} + \mathcal{O}(B \log N),
\end{equation}
so our method preserves the same model-scale dominant terms as GAD and adds only grouped scalar processing on top of the critic-emitted sequence rewards. The key practical implication is that GRGC does \emph{not} introduce an extra rollout loop, an extra critic network, or another model-scale optimization stage; its additional cost comes only from cheap per-group operations after the main neural computation has already been done.
In particular, BT loss formation, GRPO normalization, OT matching, and group power shaping are all low-order operations on scalar group rewards after the dominant neural computation is finished. Since $N$ is small and fixed in our setting, these terms do not materially affect end-to-end training time, so the overall wall-clock training cost of GRGC is expected to be nearly the same as that of GAD.
Table~\ref{tab:complexity} summarizes this asymptotic comparison.

\begin{table}[tb]
\centering
\caption{Asymptotic training complexity comparison. Here $B$ is the number of sampled student responses in a minibatch, $N$ is the prompt-group size, and $L$ is the average response length. Since $N$ is a small fixed group size in our setting, grouped operations with complexity $\mathcal{O}(B\log N)$ behave as low-order overheads and do not change the dominant model-scale training complexity. As a result, the total wall-clock training time of GRGC is expected to remain nearly identical to that of GAD.}
\small
\begin{tabular}{l|c|c|c}
\toprule
Method & Time complexity & Memory complexity & Dominant terms \\
\midrule
GAD
& \begin{tabular}[c]{@{}c@{}}
$\mathcal{O}\!\big(B\,\mathcal{F}_{\pi}^{\mathrm{roll}}(L)\big)$ \\
$+\mathcal{O}\!\big(B\,\mathcal{F}_{D}(L)\big)$ \\
$+\mathcal{O}\!\big(B\,\mathcal{F}_{\pi}^{\mathrm{upd}}(L)\big)$ \\
$+\mathcal{O}(B)$
\end{tabular}
& \begin{tabular}[c]{@{}c@{}}
$\mathcal{M}_{\pi}+\mathcal{M}_{D}$ \\
$+\mathcal{O}(B)$
\end{tabular}
& \begin{tabular}[c]{@{}l@{}}
rollout \\
critic forward/backward \\
actor update
\end{tabular}
\\
\midrule
GRGC (ours)
& \begin{tabular}[c]{@{}c@{}}
$\mathcal{O}\!\big(B\,\mathcal{F}_{\pi}^{\mathrm{roll}}(L)\big)$ \\
$+\mathcal{O}\!\big(B\,\mathcal{F}_{D}(L)\big)$ \\
$+\mathcal{O}\!\big(B\,\mathcal{F}_{\pi}^{\mathrm{upd}}(L)\big)$ \\
$+\mathcal{O}(B \log N)$
\end{tabular}
& \begin{tabular}[c]{@{}c@{}}
$\mathcal{M}_{\pi}+\mathcal{M}_{D}$ \\
$+\mathcal{O}(B)$
\end{tabular}
& \begin{tabular}[c]{@{}l@{}}
rollout + critic + actor update \\
OT sorting/matching \\
group power shaping
\end{tabular}
\\
\bottomrule
\end{tabular}
\label{tab:complexity}
\end{table}

\section{Broader Impact}
\label{a6}
Our work studies how to make black-box distillation more effective when the teacher is accessible only through output queries. On the positive side, better black-box distillation can reduce the cost of obtaining capable small language models, which may improve accessibility for research labs, educational use, and resource-constrained deployment settings. It may also enable more systematic study of proprietary-model behavior using open students that are easier to inspect, evaluate, and stress-test.

At the same time, improvements in distillation efficiency can also accelerate the transfer of undesirable behaviors from proprietary teachers into cheaper and more deployable students. More capable distilled models may be misused for low-cost spam generation, manipulative persuasion, or unsafe domain advice. In addition, more effective distillation can make capability transfer easier to reproduce, which is a dual-use concern. Our method does not require additional access to private model internals or private data, but it can increase the effectiveness of capability transfer from black-box teachers. We therefore view stronger safety evaluation, misuse screening, and downstream deployment constraints as important complements to progress in black-box distillation.

\section{Limitations}
\label{a7}
Our method is built on top of GAD and inherits both its strengths and its limitations. In particular, although GRGC improves the quality of on-policy black-box distillation, it still relies on an adversarial critic and on-policy rollouts, and is therefore typically slower than simple supervised baselines such as SeqKD. This is a real tradeoff: our goal is not to make on-policy distillation cheaper than off-policy imitation in absolute terms, but to make the extra optimization cost buy a better-conditioned and more effective training signal.

\clearpage
\newpage
\section*{NeurIPS Paper Checklist}

\begin{enumerate}

\item {\bf Claims}
    \item[] Question: Do the main claims made in the abstract and introduction accurately reflect the paper's contributions and scope?
    \item[] Answer: \answerYes{}
    \item[] Justification: The abstract and introduction state the core claim that GRGC improves on-policy black-box distillation by conditioning prompt-group reward geometry, and this matches the method, experiments, and limitations discussed in the paper; see the abstract, Section~1, and Section~5.
    \item[] Guidelines:
    \begin{itemize}
        \item The answer \answerNA{} means that the abstract and introduction do not include the claims made in the paper.
        \item The abstract and/or introduction should clearly state the claims made, including the contributions made in the paper and important assumptions and limitations. A \answerNo{} or \answerNA{} answer to this question will not be perceived well by the reviewers. 
        \item The claims made should match theoretical and experimental results, and reflect how much the results can be expected to generalize to other settings. 
        \item It is fine to include aspirational goals as motivation as long as it is clear that these goals are not attained by the paper. 
    \end{itemize}

\item {\bf Limitations}
    \item[] Question: Does the paper discuss the limitations of the work performed by the authors?
    \item[] Answer: \answerYes{}
    \item[] Justification: The paper includes an explicit Limitations discussion in the appendix, covering the additional training cost relative to SeqKD; see Appendix ``Limitations''.
    \item[] Guidelines:
    \begin{itemize}
        \item The answer \answerNA{} means that the paper has no limitation while the answer \answerNo{} means that the paper has limitations, but those are not discussed in the paper. 
        \item The authors are encouraged to create a separate ``Limitations'' section in their paper.
        \item The paper should point out any strong assumptions and how robust the results are to violations of these assumptions (e.g., independence assumptions, noiseless settings, model well-specification, asymptotic approximations only holding locally). The authors should reflect on how these assumptions might be violated in practice and what the implications would be.
        \item The authors should reflect on the scope of the claims made, e.g., if the approach was only tested on a few datasets or with a few runs. In general, empirical results often depend on implicit assumptions, which should be articulated.
        \item The authors should reflect on the factors that influence the performance of the approach. For example, a facial recognition algorithm may perform poorly when image resolution is low or images are taken in low lighting. Or a speech-to-text system might not be used reliably to provide closed captions for online lectures because it fails to handle technical jargon.
        \item The authors should discuss the computational efficiency of the proposed algorithms and how they scale with dataset size.
        \item If applicable, the authors should discuss possible limitations of their approach to address problems of privacy and fairness.
        \item While the authors might fear that complete honesty about limitations might be used by reviewers as grounds for rejection, a worse outcome might be that reviewers discover limitations that aren't acknowledged in the paper. The authors should use their best judgment and recognize that individual actions in favor of transparency play an important role in developing norms that preserve the integrity of the community. Reviewers will be specifically instructed to not penalize honesty concerning limitations.
    \end{itemize}

\item {\bf Theory assumptions and proofs}
    \item[] Question: For each theoretical result, does the paper provide the full set of assumptions and a complete (and correct) proof?
    \item[] Answer: \answerYes{}
    \item[] Justification: The appendix provides explicit assumptions, formal propositions, and corresponding derivations and proof sketches for the theoretical claims used in the paper; see Appendix ``Mechanism Analysis for OT Calibration and Group Power''.
    \item[] Guidelines:
    \begin{itemize}
        \item The answer \answerNA{} means that the paper does not include theoretical results. 
        \item All the theorems, formulas, and proofs in the paper should be numbered and cross-referenced.
        \item All assumptions should be clearly stated or referenced in the statement of any theorems.
        \item The proofs can either appear in the main paper or the supplemental material, but if they appear in the supplemental material, the authors are encouraged to provide a short proof sketch to provide intuition. 
        \item Inversely, any informal proof provided in the core of the paper should be complemented by formal proofs provided in appendix or supplemental material.
        \item Theorems and Lemmas that the proof relies upon should be properly referenced. 
    \end{itemize}

    \item {\bf Experimental result reproducibility}
    \item[] Question: Does the paper fully disclose all the information needed to reproduce the main experimental results of the paper to the extent that it affects the main claims and/or conclusions of the paper (regardless of whether the code and data are provided or not)?
    \item[] Answer: \answerYes{}
    \item[] Justification: The paper specifies datasets, model families, teacher models, evaluation benchmarks, key hyperparameters, training schedule, and hardware used for the reported experiments; see Section~4.1 and the appendix.
    \item[] Guidelines:
    \begin{itemize}
        \item The answer \answerNA{} means that the paper does not include experiments.
        \item If the paper includes experiments, a \answerNo{} answer to this question will not be perceived well by the reviewers: Making the paper reproducible is important, regardless of whether the code and data are provided or not.
        \item If the contribution is a dataset and\slash or model, the authors should describe the steps taken to make their results reproducible or verifiable. 
        \item Depending on the contribution, reproducibility can be accomplished in various ways. For example, if the contribution is a novel architecture, describing the architecture fully might suffice, or if the contribution is a specific model and empirical evaluation, it may be necessary to either make it possible for others to replicate the model with the same dataset, or provide access to the model. In general. releasing code and data is often one good way to accomplish this, but reproducibility can also be provided via detailed instructions for how to replicate the results, access to a hosted model (e.g., in the case of a large language model), releasing of a model checkpoint, or other means that are appropriate to the research performed.
        \item While NeurIPS does not require releasing code, the conference does require all submissions to provide some reasonable avenue for reproducibility, which may depend on the nature of the contribution. For example
        \begin{enumerate}
            \item If the contribution is primarily a new algorithm, the paper should make it clear how to reproduce that algorithm.
            \item If the contribution is primarily a new model architecture, the paper should describe the architecture clearly and fully.
            \item If the contribution is a new model (e.g., a large language model), then there should either be a way to access this model for reproducing the results or a way to reproduce the model (e.g., with an open-source dataset or instructions for how to construct the dataset).
            \item We recognize that reproducibility may be tricky in some cases, in which case authors are welcome to describe the particular way they provide for reproducibility. In the case of closed-source models, it may be that access to the model is limited in some way (e.g., to registered users), but it should be possible for other researchers to have some path to reproducing or verifying the results.
        \end{enumerate}
    \end{itemize}

\item {\bf Open access to data and code}
    \item[] Question: Does the paper provide open access to the data and code, with sufficient instructions to faithfully reproduce the main experimental results, as described in supplemental material?
    \item[] Answer: \answerYes{}
    \item[] Justification: The paper and appendix provide the datasets, model families, evaluation prompts, pseudocode, hyperparameters, and reproduction-relevant implementation details needed to reproduce the main experiments.
    \item[] Guidelines:
    \begin{itemize}
        \item The answer \answerNA{} means that paper does not include experiments requiring code.
        \item Please see the NeurIPS code and data submission guidelines (\url{https://neurips.cc/public/guides/CodeSubmissionPolicy}) for more details.
        \item While we encourage the release of code and data, we understand that this might not be possible, so \answerNo{} is an acceptable answer. Papers cannot be rejected simply for not including code, unless this is central to the contribution (e.g., for a new open-source benchmark).
        \item The instructions should contain the exact command and environment needed to run to reproduce the results. See the NeurIPS code and data submission guidelines (\url{https://neurips.cc/public/guides/CodeSubmissionPolicy}) for more details.
        \item The authors should provide instructions on data access and preparation, including how to access the raw data, preprocessed data, intermediate data, and generated data, etc.
        \item The authors should provide scripts to reproduce all experimental results for the new proposed method and baselines. If only a subset of experiments are reproducible, they should state which ones are omitted from the script and why.
        \item At submission time, to preserve anonymity, the authors should release anonymized versions (if applicable).
        \item Providing as much information as possible in supplemental material (appended to the paper) is recommended, but including URLs to data and code is permitted.
    \end{itemize}

\item {\bf Experimental setting/details}
    \item[] Question: Does the paper specify all the training and test details (e.g., data splits, hyperparameters, how they were chosen, type of optimizer) necessary to understand the results?
    \item[] Answer: \answerYes{}
    \item[] Justification: Section~4.1 and the appendix describe datasets, evaluation splits, teacher and student models, baselines, training schedule, group size, learning rates, OT and power hyperparameters, sensitivity analyses for the two method-specific hyperparameters, and hardware.
    \item[] Guidelines:
    \begin{itemize}
        \item The answer \answerNA{} means that the paper does not include experiments.
        \item The experimental setting should be presented in the core of the paper to a level of detail that is necessary to appreciate the results and make sense of them.
        \item The full details can be provided either with the code, in appendix, or as supplemental material.
    \end{itemize}

\item {\bf Experiment statistical significance}
    \item[] Question: Does the paper report error bars suitably and correctly defined or other appropriate information about the statistical significance of the experiments?
    \item[] Answer: \answerYes{}
   \item[] Justification: The paper reports multi-seed automatic evaluation results with mean and standard deviation across five evaluation seeds in the Appendix, showing that the gains remain much larger than judge-sampling variability. It also supports the empirical conclusions through consistent trends across multiple teacher-student pairs, multiple test sets, sensitivity analyses, training-stability analyses, and complementary human evaluation.
    \item[] Guidelines:
    \begin{itemize}
        \item The answer \answerNA{} means that the paper does not include experiments.
        \item The authors should answer \answerYes{} if the results are accompanied by error bars, confidence intervals, or statistical significance tests, at least for the experiments that support the main claims of the paper.
        \item The factors of variability that the error bars are capturing should be clearly stated (for example, train/test split, initialization, random drawing of some parameter, or overall run with given experimental conditions).
        \item The method for calculating the error bars should be explained (closed form formula, call to a library function, bootstrap, etc.)
        \item The assumptions made should be given (e.g., Normally distributed errors).
        \item It should be clear whether the error bar is the standard deviation or the standard error of the mean.
        \item It is OK to report 1-sigma error bars, but one should state it. The authors should preferably report a 2-sigma error bar than state that they have a 96\% CI, if the hypothesis of Normality of errors is not verified.
        \item For asymmetric distributions, the authors should be careful not to show in tables or figures symmetric error bars that would yield results that are out of range (e.g., negative error rates).
        \item If error bars are reported in tables or plots, the authors should explain in the text how they were calculated and reference the corresponding figures or tables in the text.
    \end{itemize}

\item {\bf Experiments compute resources}
    \item[] Question: For each experiment, does the paper provide sufficient information on the computer resources (type of compute workers, memory, time of execution) needed to reproduce the experiments?
    \item[] Answer: \answerYes{}
    \item[] Justification: The paper specifies the accelerator type and count, the training schedule in epochs and optimization steps, the key sequence lengths, and the overall training recipe used for the reported experiments; see Section~4.1 and Appendix ``Complexity Analysis''.
    \item[] Guidelines:
    \begin{itemize}
        \item The answer \answerNA{} means that the paper does not include experiments.
        \item The paper should indicate the type of compute workers CPU or GPU, internal cluster, or cloud provider, including relevant memory and storage.
        \item The paper should provide the amount of compute required for each of the individual experimental runs as well as estimate the total compute. 
        \item The paper should disclose whether the full research project required more compute than the experiments reported in the paper (e.g., preliminary or failed experiments that didn't make it into the paper). 
    \end{itemize}
    
\item {\bf Code of ethics}
    \item[] Question: Does the research conducted in the paper conform, in every respect, with the NeurIPS Code of Ethics \url{https://neurips.cc/public/EthicsGuidelines}?
    \item[] Answer: \answerYes{}
    \item[] Justification: To the best of our knowledge, the research conforms to the NeurIPS Code of Ethics; the paper also discusses broader impacts and limitations, including dual-use concerns.
    \item[] Guidelines:
    \begin{itemize}
        \item The answer \answerNA{} means that the authors have not reviewed the NeurIPS Code of Ethics.
        \item If the authors answer \answerNo, they should explain the special circumstances that require a deviation from the Code of Ethics.
        \item The authors should make sure to preserve anonymity (e.g., if there is a special consideration due to laws or regulations in their jurisdiction).
    \end{itemize}

\item {\bf Broader impacts}
    \item[] Question: Does the paper discuss both potential positive societal impacts and negative societal impacts of the work performed?
    \item[] Answer: \answerYes{}
    \item[] Justification: The paper discusses both positive impacts (cheaper, more accessible small models) and negative impacts (more effective transfer of harmful behaviors), and also mentions the need for safeguards; see the main-paper Broader Impact paragraph and the appendix Broader Impact section.
    \item[] Guidelines:
    \begin{itemize}
        \item The answer \answerNA{} means that there is no societal impact of the work performed.
        \item If the authors answer \answerNA{} or \answerNo, they should explain why their work has no societal impact or why the paper does not address societal impact.
        \item Examples of negative societal impacts include potential malicious or unintended uses (e.g., disinformation, generating fake profiles, surveillance), fairness considerations (e.g., deployment of technologies that could make decisions that unfairly impact specific groups), privacy considerations, and security considerations.
        \item The conference expects that many papers will be foundational research and not tied to particular applications, let alone deployments. However, if there is a direct path to any negative applications, the authors should point it out. For example, it is legitimate to point out that an improvement in the quality of generative models could be used to generate Deepfakes for disinformation. On the other hand, it is not needed to point out that a generic algorithm for optimizing neural networks could enable people to train models that generate Deepfakes faster.
        \item The authors should consider possible harms that could arise when the technology is being used as intended and functioning correctly, harms that could arise when the technology is being used as intended but gives incorrect results, and harms following from (intentional or unintentional) misuse of the technology.
        \item If there are negative societal impacts, the authors could also discuss possible mitigation strategies (e.g., gated release of models, providing defenses in addition to attacks, mechanisms for monitoring misuse, mechanisms to monitor how a system learns from feedback over time, improving the efficiency and accessibility of ML).
    \end{itemize}
    
\item {\bf Safeguards}
    \item[] Question: Does the paper describe safeguards that have been put in place for responsible release of data or models that have a high risk for misuse (e.g., pre-trained language models, image generators, or scraped datasets)?
    \item[] Answer: \answerNA{}
    \item[] Justification: This submission does not release a new high-risk model, model checkpoint, or scraped dataset asset; the paper instead discusses broader-impact risks and deployment considerations.
    \item[] Guidelines:
    \begin{itemize}
        \item The answer \answerNA{} means that the paper poses no such risks.
        \item Released models that have a high risk for misuse or dual-use should be released with necessary safeguards to allow for controlled use of the model, for example by requiring that users adhere to usage guidelines or restrictions to access the model or implementing safety filters. 
        \item Datasets that have been scraped from the Internet could pose safety risks. The authors should describe how they avoided releasing unsafe images.
        \item We recognize that providing effective safeguards is challenging, and many papers do not require this, but we encourage authors to take this into account and make a best faith effort.
    \end{itemize}

\item {\bf Licenses for existing assets}
    \item[] Question: Are the creators or original owners of assets (e.g., code, data, models), used in the paper, properly credited and are the license and terms of use explicitly mentioned and properly respected?
    \item[] Answer: \answerYes{}
    \item[] Justification: The paper credits the datasets, models, and baseline methods used in the experiments and identifies the main external assets explicitly in the experimental sections and references.
    \item[] Guidelines:
    \begin{itemize}
        \item The answer \answerNA{} means that the paper does not use existing assets.
        \item The authors should cite the original paper that produced the code package or dataset.
        \item The authors should state which version of the asset is used and, if possible, include a URL.
        \item The name of the license (e.g., CC-BY 4.0) should be included for each asset.
        \item For scraped data from a particular source (e.g., website), the copyright and terms of service of that source should be provided.
        \item If assets are released, the license, copyright information, and terms of use in the package should be provided. For popular datasets, \url{paperswithcode.com/datasets} has curated licenses for some datasets. Their licensing guide can help determine the license of a dataset.
        \item For existing datasets that are re-packaged, both the original license and the license of the derived asset (if it has changed) should be provided.
        \item If this information is not available online, the authors are encouraged to reach out to the asset's creators.
    \end{itemize}

\item {\bf New assets}
    \item[] Question: Are new assets introduced in the paper well documented and is the documentation provided alongside the assets?
    \item[] Answer: \answerNA{}
    \item[] Justification: The current submission does not release a new dataset, benchmark, model checkpoint, or public code package as an asset.
    \item[] Guidelines:
    \begin{itemize}
        \item The answer \answerNA{} means that the paper does not release new assets.
        \item Researchers should communicate the details of the dataset\slash code\slash model as part of their submissions via structured templates. This includes details about training, license, limitations, etc. 
        \item The paper should discuss whether and how consent was obtained from people whose asset is used.
        \item At submission time, remember to anonymize your assets (if applicable). You can either create an anonymized URL or include an anonymized zip file.
    \end{itemize}

\item {\bf Crowdsourcing and research with human subjects}
    \item[] Question: For crowdsourcing experiments and research with human subjects, does the paper include the full text of instructions given to participants and screenshots, if applicable, as well as details about compensation (if any)? 
    \item[] Answer: \answerYes{}
    \item[] Justification: The appendix describes the human-evaluation setup, the annotator pool, the per-model sample allocation, and the evaluation criteria used by the annotators. The study does not rely on an external crowdsourcing platform.
    \item[] Guidelines:
    \begin{itemize}
        \item The answer \answerNA{} means that the paper does not involve crowdsourcing nor research with human subjects.
        \item Including this information in the supplemental material is fine, but if the main contribution of the paper involves human subjects, then as much detail as possible should be included in the main paper. 
        \item According to the NeurIPS Code of Ethics, workers involved in data collection, curation, or other labor should be paid at least the minimum wage in the country of the data collector. 
    \end{itemize}

\item {\bf Institutional review board (IRB) approvals or equivalent for research with human subjects}
    \item[] Question: Does the paper describe potential risks incurred by study participants, whether such risks were disclosed to the subjects, and whether Institutional Review Board (IRB) approvals (or an equivalent approval/review based on the requirements of your country or institution) were obtained?
    \item[] Answer: \answerYes{}
    \item[] Justification: The human evaluation concerns expert assessment of model outputs only, does not involve collection of personal or sensitive participant data, and is described in the appendix together with the evaluation procedure and scope of human involvement.
    \item[] Guidelines:
    \begin{itemize}
        \item The answer \answerNA{} means that the paper does not involve crowdsourcing nor research with human subjects.
        \item Depending on the country in which research is conducted, IRB approval (or equivalent) may be required for any human subjects research. If you obtained IRB approval, you should clearly state this in the paper. 
        \item We recognize that the procedures for this may vary significantly between institutions and locations, and we expect authors to adhere to the NeurIPS Code of Ethics and the guidelines for their institution. 
        \item For initial submissions, do not include any information that would break anonymity (if applicable), such as the institution conducting the review.
    \end{itemize}

\item {\bf Declaration of LLM usage}
    \item[] Question: Does the paper describe the usage of LLMs if it is an important, original, or non-standard component of the core methods in this research? Note that if the LLM is used only for writing, editing, or formatting purposes and does \emph{not} impact the core methodology, scientific rigor, or originality of the research, declaration is not required.
    \item[] Answer: \answerNA{}
    \item[] Justification: The core contribution is a reward-geometry conditioning method for black-box distillation rather than a method that uses an LLM as a novel algorithmic component; teacher and judge models are standard experimental ingredients and are documented in Section~4.1.
    \item[] Guidelines:
    \begin{itemize}
        \item The answer \answerNA{} means that the core method development in this research does not involve LLMs as any important, original, or non-standard components.
        \item Please refer to our LLM policy in the NeurIPS handbook for what should or should not be described.
    \end{itemize}

\end{enumerate}

\end{document}